\documentclass[journal,twoside,web]{ieeecolor}
\usepackage{generic}

\usepackage{algorithmic}
\usepackage{textcomp}

\usepackage{import} 
\usepackage[pdftex]{graphicx}
\usepackage{cite}
\usepackage{verbatim}		
\usepackage{amsmath}
\usepackage{amssymb}

\usepackage{amsthm}
\usepackage{mathtools}	
\usepackage{grffile}	
\usepackage[tight,footnotesize]{subfigure}
\usepackage{microtype} 
\usepackage{color}
\usepackage{url}
\usepackage[ruled,vlined,linesnumbered]{algorithm2e}
\usepackage{bm} 
\usepackage{comment}
\usepackage{arydshln}
\let\labelindent\relax
\usepackage{enumitem}

\usepackage{adjustbox}
\usepackage{tikz}
\usetikzlibrary{shadings, patterns, angles, quotes, arrows.meta, shapes, decorations.pathmorphing, decorations.shapes, decorations.text, positioning}
\usetikzlibrary{calc,intersections,arrows.meta}
\usepackage{pgfplots}

\usepackage{hyperref}
\hypersetup{
	colorlinks=true,
	linkcolor=blue,
	citecolor=red,
}

	\tikzset{
	pil/.style={
		->,
		thick,
		shorten <=2pt,
		shorten >=2pt,}
}

\theoremstyle{remark}
\newtheorem{remarkx}{Remark}
\newenvironment{remark}
{\pushQED{\qed}\remarkx}
{\popQED\endremarkx}
\newtheorem{examplex}{Example}
\newenvironment{example}
{\pushQED{\qed}\examplex}
{\popQED\endexamplex}

\theoremstyle{definition}
\newtheorem{defn}{Definition}

\newtheorem{problem}{Problem}
\newtheorem*{problem*}{Problem}

\theoremstyle{plain}
\newtheorem{theorem}{Theorem}
\newtheorem{lemma}{Lemma}
\newtheorem{coroll}{Corollary}
\newtheorem{prop}{Proposition}

\newcommand{\cmmnt}[1]{}

\newcommand{\norm}[1]{\left\lVert#1\right\rVert}

\newcommand{\scalemath}[2]{\scalebox{#1}{\mbox{\ensuremath{\displaystyle #2}}}}

\tikzstyle{block} = [draw, fill=gray!30, rectangle, 
    minimum height=3em, minimum width=3em, text width=3cm, align=center]
\tikzstyle{block_red} = [draw, fill=gray!30, rectangle, 
    minimum height=3em, minimum width=3em, text width=3cm, align=center]
\tikzstyle{block_small} = [draw, fill=gray!30, rectangle, 
    minimum height=2em, minimum width=2em, text width=1cm, align=center]
\tikzstyle{block_kin} = [draw, fill=gray!30, rectangle, 
    minimum height=3em, minimum width=3em, text width=3cm, align=center]
\tikzstyle{block_dashed} = [draw, dashed, line width=0.5mm, fill=gray!30, rectangle, 
    minimum height=3em, minimum width=7cm, minimum height=1cm, text width=6cm, align=center]
\tikzstyle{block_nei} = [draw, dashed, line width=0.5mm, fill=gray!30, rectangle, 
    minimum height=3em, minimum width=3cm, minimum height=1cm, text width=3cm, align=center]

\tikzstyle{point} = [coordinate]

\title{\LARGE \bf
Fully distributed and resilient source seeking for robot swarms
}

\author{Jesus Bautista*, Antonio Acuaviva*, Jose Hinojosa, Weijia Yao, \\ Juan Jimenez, \IEEEmembership{Member, IEEE}, Hector Garcia de Marina, \IEEEmembership{Member, IEEE} 
\thanks{Jesus Bautista, Jose Hinojosa, and H. Garcia de Marina are with the Department of Computer Engineering, Automation, and Robotics, and the Institute of Mathematics (IMAG) at the University of Granada, Spain. A. Acuaviva is with the School of Mathematical Sciences, University of Lancaster, United Kingdom. J. Jimenez is with the Department of Computer Architecture and Automation, Universidad Complutense de Madrid, Spain.  Weijia Yao is with the School of Robotics, Hunan University. Corresponding author e-mail: {\tt\small hgdemarina@ugr.es}. The work of H.G. de Marina is supported by the grant Ramon y Cajal RYC2020-030090-I from the Spanish Ministry of Science, and the ERC Starting Grant \emph{iSwarm} 101076091. * Bautista and Acuaviva contributed equally as first authors in this publication.}%
}

\begin{document}

\maketitle
\thispagestyle{empty}
\pagestyle{empty}

\begin{abstract}
	Existing source-seeking algorithms for robot swarms typically require either direct gradient measurements or rigid geometric formations, limiting their flexibility and resilience to robot failures. We propose a fully distributed solution that overcomes these limitations by computing an ascending direction through local field measurements and distributed estimation of centroid-relative coordinates. The resulting architecture consists of three exponentially convergent algorithms operating in a slow-fast closed-loop system, enabling simultaneous estimation and motion control without central coordination. Our framework accommodates arbitrary swarm geometries and analyzes how the spatial distribution of robots affects gradient observability, robustness, and resilience to failures. We characterize optimal swarm shapes that guarantee alignment with the true gradient and show how shape morphing can maneuver the collective motion. The approach is developed for kinematic points in $\mathbb{R}^m$ and extended to 2D unicycles with constant speed. Simulations with large-scale swarms validate the methodology.
\end{abstract}


\section{Introduction}

\subsection{Motivation and aim}
The ability to detect and surround sources of chemicals, pollution, and radio signals effectively with robot swarms can enable persistent missions in vast areas for environmental monitoring, search \& rescue, and precision agriculture operations \cite{ogren2004cooperative, kumar2004robot, mcguire2019minimal, li2006moth,twigg2012rss}. In particular, seeking the source of a scalar field can be regarded as one fundamental task in swarm robotics \cite{yang2018grand,brambilla2013swarm} where the resiliency of the robot swarm is expected, i.e., the swarm preserves its functionality against unexpected adverse conditions, unknown (possibly unmodeled) disturbances and the malfunctioning of robot individuals. However, formal guarantees on the performance of a solution while remaining feasible in practice, e.g., by fitting into realistic robot dynamics, communications, and scalability, are one of the great challenges in swarm robotics \cite{yang2018grand,dorigo2021swarm,brambilla2013swarm}. In this regard, we present and rigorously analyze a distributed source-seeking strategy for robot swarms, where the source is reached even if individuals go missing during the mission.

\subsection{Overview of our source-seeking strategy}
To facilitate a fair comparison with existing algorithms in the next subsection, we first outline the key features of our solution. Our strategy guarantees with sufficient conditions an ascending direction rather than the gradient in order to guide the centroid of the robot swarm to the source of a scalar field, i.e., all the individual robots estimate distributively and track a common ascending direction. Indeed, considering an ascending direction offers more flexibility to the robot swarm since there are multiple ascending directions instead of the single gradient direction. The algorithm does not require a specific formation or shape for the robot swarm; in fact, we will show how the computed ascending direction reacts to the relation between the swarm shape and the scalar field. This fact will be exploited to maneuver the robot swarm by morphing its shape while getting closer to the source simultaneously. The lack of a shape requirement allows the algorithm to maintain performance even if some robots are missing, as the resulting formation shape still follows an ascending direction.

The proposed mathematical formulation for computing the ascending direction enables the extension from discrete robot configurations to a continuous distribution of robots. Indeed, the analysis of the swarm shape and the density of the robot distribution within allows us to guarantee \emph{optimal shapes} so that the calculated ascending direction is parallel to the actual gradient. We will show that the ascending direction can be calculated distributively based on a consensus algorithm. Finally, since robots need it for the ascending direction estimation, we analyze a distributed algorithm to estimate the swarm centroid.

The presented solution is fully compatible with single integrator robots, and with the restrictive but more realistic unicycle (i.e., Dubins car model) robots with a constant speed. Indeed, inspired by the analysis on how unicycles can track guiding vector fields \cite{yao2021singularity,yao2022guiding}, we show how a team of 2D unicycles travelling at a constant speed can track a vector field consisting of ascending directions, and how the centroid of the team gets arbitrarily close to the source eventually. The nature of our swarm algorithm is purely reactive to the field, and it does not need scouting or gathering prior information about the scalar field, and it does not consider a bounded map either. In that regards, our algorithm is mainly suited to work in scalar fields with only one maximum point and whose gradient and Hessian are bounded.

\subsection{Existing solutions: comparisons and trade-offs}
This subsection reviews the advantages and limitations of source-seeking algorithms commonly employed in the literature. We keep in mind that robot systems are tailored to specific missions and requirements, e.g., robot dynamics, sensing restrictions, traveled distance, or even prior knowledge of the scenario, i.e., it is fair to acknowledge that there is no better algorithm than others in every single metric regarding a source-seeking task.

The popular field-climbing methods focus on estimating the gradient, and possibly the Hessian as well, of an unknown scalar field. In \cite{rosero2014cooperative, barogh2017cooperative}, the gradient estimation occurs in a distributed manner, where each robot calculates the different gradients at their positions. Since each local gradient is different, a common direction for the team is achieved through a distance-based formation controller that maintains the cohesion of the robot swarm. It is worth noting that this method fails when a subset of neighboring robots adopts a \emph{degenerate} shape, e.g., a line in a 2D plane. In addition, the formation shape and following the gradient in \cite{rosero2014cooperative, barogh2017cooperative} are not fully compatible tasks and they disturb each other, affecting the mission performance. Alternatively, in \cite{brinon2015distributed, brinon2019multirobot}, the gradient and Hessian are estimated at the centroid of a circular formation consisting of unicycles. However, this methodology is relatively rigid due to its mandatory circular shape. Such a formation of unicycles is also presented in \cite{fabbiano2016distributed}, where robots utilize relative heading measurements rather than relative position measurements. In contrast, our algorithm calculates an ascending direction instead of the gradient, allowing our robot team to enjoy greater flexibility concerning formation shapes while still offering formal guarantees for reaching the source. Finally, for gradient-following methods under known field assumptions, a pioneering work is established in \cite{ogren2004cooperative}.

Extremum seeking is another technique employed extensively as discussed in \cite{li2020cooperative, cochran2009nonholonomic, biyik2008gradient}. In these studies, robots, which could potentially be just one, with nonholonomic dynamics execute periodic motions to facilitate a gradient estimation. However, when multiple robots are involved, there is a necessity for the exchange of estimated parameters. Furthermore, the resulting robot trajectories are often characterized by long and tortuous paths with sharp turns. In contrast, our algorithm generates smoother trajectories suitable for robots such as fixed-wing drones. 

All the algorithms above require communication among robots sharing the measured strength of the scalar field. Nonetheless, the authors in \cite{al2021distributed} offer an elegant solution involving the principal component analysis of the field that requires no sharing of the field strength. However, the centroid of the swarm is still necessary and communication might be in order to estimate it. It is worth noting that in \cite{al2021distributed}, the initial positions of the robots determine the (time-varying) formation during the mission, and it is not under control. In comparison, although our algorithm requires communication among robots due to the usage of a consensus algorithm, it is compatible with formation control laws, at least for single-integrators, so that it makes the trajectory of the robots more predictable.

Another family of source-seeking algorithms is information-based, which differs from the previously discussed algorithms, including our proposed one, as they are not purely reactive to the field. Information-based strategies can handle multiple sources but at the cost of requiring prior knowledge of the field, or at least they necessitate additional stages before seeking in order to estimate or scout the scalar field. However, the performance of these algorithms is more often supported by compelling experimental evidence rather than analytical proofs \cite{zhang2023distributed,bayat2016optimal,lee2018active}. In connection with information-based algorithms, we conclude this review by briefly discussing a related problem, namely, the mobile-sensing coverage problem \cite{cortes2004coverage, cortes2009, benevento2020multi}. Algorithms addressing this problem are particularly effective at handling multiple sources simultaneously with a robot swarm. They primarily rely on partitioning the area using Voronoi cells. However, the performance comes at a higher cost compared to typical source-seeking solutions: individual robots require dense or continuous information about the scalar field within their corresponding Voronoi cell, and the area to be covered must be predetermined.

\subsection{Extensions from our previous conference paper}
Compared to our conference contribution \cite{acuaviva2023resilient}, this journal paper provides several substantial theoretical and architectural advances:
\begin{itemize}
	\item \textbf{Structural analysis of the ascent direction.}  
	We provide a complete characterization of the distributed ascent direction and show how its alignment with the true gradient depends on the spatial variance of the swarm. This unifies previously treated geometric cases and reveals the fundamental role played by the swarm's shape.
	\item \textbf{Role of symmetry and higher-order effects.}  
	We show that centrally symmetric deployments eliminate curvature-induced errors in the ascent direction. This establishes precise conditions under which the swarm behaves as an accurate gradient sensor and explains why certain formations are inherently more robust.
	\item \textbf{Swarm morphing and maneuverability.}  
	We demonstrate how anisotropic scaling of the formation modifies the directional bias of the ascent direction. This provides a geometric mechanism to maneuver the swarm by reshaping its spatial distribution.
	\item \textbf{Quantified resilience to failures.}  
	We formally relate robustness to the swarm's spatial dispersion and show that the impact of losing individual robots decreases as the swarm size increases. This provides a rigorous explanation of resilience in large-scale robot teams.
	\item \textbf{Fully distributed implementation.}  
	Unlike the conference version, which relied on centralized computation and complete communication, we develop a leaderless distributed architecture that only requires network connectivity. This includes a distributed estimation of the swarm centroid and a distributed computation of the ascent direction.
	\item \textbf{Slow-fast closed-loop integration and unicycle extension.}  
	We prove exponential convergence of the distributed estimators and tracking controller, enabling their integration in a slow-fast closed-loop system. The analysis is extended from kinematic point robots in Euclidean spaces to 2D-unicycle robots traveling at constant speed, ensuring that the swarm approaches the source with arbitrary precision.
\end{itemize}

\subsection{Organization}
The article is organized as follows. Section \ref{sec: pre} introduces the notation, robot models, and problem formulation. Section \ref{sec: tools} develops the analytical tools for computing an ascending direction and analyzes how swarm geometry affects robustness. Section \ref{sec: arch} presents the distributed architecture, including algorithms for centroid estimation and ascending direction computation. Section \ref{sec: uni} extends the analysis to constant-speed 2D unicycle robots within a slow-fast closed-loop framework. Section \ref{sec: sce} provides simulation results, and Section \ref{sec: con} concludes the paper.

\section{Preliminaries and problem formulation}
\label{sec: pre}

\subsection{Generic notation}
Given a matrix $A\in\mathbb{R}^{p\times q}$, we define the operator $\overline A := A \otimes I_m \in \mathbb{R}^{pm \times qm}$, where $\otimes$ denotes the Kronecker product and $I_m$ is the identity matrix with dimension $m = \{2,3\}$. We denote by $\mathbf{1}_p\in\mathbb{R}^p$ the all-one column vector; by $\|\cdot\|$, the standard Euclidean norm for vectors and the induced 2-norm for matrices; by $A^\top$ the transpose of the vector/matrix $A$; and by $|A|$ the determinant of $A$. 
The $+\pi/2$ rotation matrix is given by $E = \left[ \begin{smallmatrix} 0 & -1 \\ 1 & 0 \end{smallmatrix} \right]$. 
Finally, $\text{O}(m)$ denotes the full orthogonal group (rotations and reflections), $\text{SO}(m)$ the special orthogonal group (rotations only), and the $n$-sphere is defined as $\mathbb{S}^n := \{x \in \mathbb{R}^{n+1}\, | \, \|x\| = 1\}$. In the case of the 1-sphere, we use the angular representation $\mathbb{S}^1 \simeq \mathbb{T}$, where $\mathbb{T} := (-\pi, \pi]$.

\subsection{Graph Theory}
Since our proposed source-seeking solution consists of distributed algorithms, we need to introduce some notions from Graph Theory \cite{bullo2020lectures} in order to precisely define the relationships between the robots. Consider a group of $N > m$ robots, then a \emph{graph} $\mathcal{G} = (\mathcal{V}, \mathcal{E})$ consists of two non-empty sets: the node set $\mathcal{V} := \{1,\dots,N\}$ where each node $i$ corresponds to the robot $i$, and the ordered edge set $\mathcal{E} \subseteq (\mathcal{V}\times\mathcal{V})$ defining the communications or sensing between pairs of different robots. For an arbitrary edge $\mathcal{E}_k = (\mathcal{E}_k^{\text{head}},\mathcal{E}_k^{\text{tail}})$, we call its first and second element the \emph{head} and the \emph{tail} respectively. The set $\mathcal{N}_i$ containing the neighbors of the node $i$ is defined by $\mathcal{N}_i:=\{j\in\mathcal{V}:(i,j)\in\mathcal{E}\}$. 

We only deal with the special case of \emph{undirected} graphs throughout the article, where all the edges $\mathcal{E}_k$ are considered \emph{bidirectional}, i.e., if $(i,j)\in\mathcal{E}$ then it implies that $(j,i)\in\mathcal{E}$. For an undirected graph, we choose only one of these two arbitrary directions between nodes $i$ and $j$, so that we can construct the \emph{incidence matrix} $B\in\mathbb{R}^{|\mathcal{V}|\times |\mathcal{E}|}$ of $\mathcal{G}$ as $b_{ik} = 1$ if $i = \mathcal{E}_k^{\text{tail}}$, $-1$ if $i = \mathcal{E}_k^{\text{head}}$ and $0$ otherwise. For an undirected graph, the \emph{Laplacian matrix} $L\in\mathbb{R}^{|\mathcal{V}|\times |\mathcal{V}|}$ \cite[Chapter 6]{bullo2020lectures} can be calculated as $L = BB^\top.$ If the graph $\mathcal{G}$ is \emph{connected}\cite[Chapter 3]{bullo2020lectures}, then the Laplacian matrix $L$ has a single eigenvalue equal to zero, whose associated eigenvector is $\mathbf{1}_N$ since $B^\top\mathbf{1}_N = 0$; thus, $L$ is positive semidefinite.

\subsection{Robot dynamics and deployment}
The position of the $i \in \mathcal{V}$ robot in the Euclidean space is represented by $p_i \in \mathbb{R}^m$. We define $p\in\mathbb{R}^{mN}$ as the stacked vector of the positions of all robots in the team, and the centroid of the team as $p_c := \frac{1}{N}\sum_{i=1}^{N}p_i$. Thus, we can write $p_i = p_c + x_i$, where $x_i\in\mathbb{R}^m$ for all $i\in\mathcal{V}$ represents the centroid-relative coordinates, describing how the robots are distributed around the centroid.

In this work, we consider both dynamics for the robots, the single integrator and the 2D unicycle with constant speed. The single integrator dynamics are modeled by
\begin{equation}
	\label{eq: si}
\dot p_i = u_i,
\end{equation}
where $u_i \in\mathbb{R}^m$ is the control action, or guidance velocity in practice.
The 2D unicycle model, assuming without loss of generality a constant speed of $1$ in a convenient unit system, is given by 
\begin{equation} \label{eq: ud}
	\begin{cases}
	\dot p_i = \left[\cos{\alpha_i} \; \sin{\alpha_i}\right]^\top \\
	\dot\alpha_i = \omega_i,
	\end{cases}
\end{equation}
where $\omega_i\in\mathbb{R}$ is the angular velocity, serving as the control input that drives the robot's heading angle $\alpha_i\in \mathbb{T}$.

For a robot swarm, we define its spatial arrangement, referred to as \emph{deployment}, as follows.
\begin{defn}
The \emph{deployment} \label{deployment} of a robot team is defined as the stacked vector $x := \left[x_1^\top \cdots \, x_N^\top\right]^\top \in \mathbb{R}^{mN}$ satisfying $\sum_{i=1}^N x_i = 0$. We say that a deployment is not \emph{degenerated} if all $x_i$'s span $\mathbb{R}^m$.
\end{defn}
Therefore, a necessary condition for a deployment not being degenerated is that $N > m$. Additionally, a deployment is degenerated if and only if $\operatorname{rank}(P(x)) < m$, where $P(x) := \frac{1}{N}\sum_{i=1}^N x_i x_i^\top$ is the covariance matrix of the deployment.

\subsection{The scalar field and problem statement}
The strength of a signal throughout the space can be described by a scalar field.

\begin{defn} \label{signal} 
    A \emph{signal} is a scalar field $\sigma: \mathbb{R}^m \to \mathbb{R}^+$ which is class $C^2$ (twice continuously differentiable) and all its partial derivatives up to second order are bounded uniformly. Also, $\sigma$ has only one maximum at $p_\sigma \in \mathbb{R}^m$, i.e., the source of the signal, and its gradient at $a\in\mathbb{R}^m$ satisfies $\nabla\sigma(a) \neq 0 \iff a \neq p_\sigma$ and $\lim_{|a|\to\infty}\sigma(a) = 0$.
\end{defn}

Our definition of signal encompasses numerous physical-world models, especially Gaussian distributed signals and signals that decay from the source following a power law $x^{-\alpha}$ (where $2 \leq \alpha \leq 3$) beyond a minimum distance $x_{min}$ from the origin \cite{clauset2009power}. Such signal strengths can represent the modulus of the electromagnetic field, the concentration of a contaminant, or heat radiation. In particular, quadratic scalar fields are relevant in the physical world when considering the inverse of the power law $x^{-2}$ or the logarithm of a Gaussian distribution.

In this paper, the gradient is defined as a column vector, i.e., $\nabla\sigma(\cdot) \in\mathbb{R}^m$, and according to our definition of a signal:
\begin{equation} \label{eq_grad_hess_bound}
\|\nabla\sigma(a)\| \leq K \quad \text{and} \quad \|H_\sigma(a)\| \leq 2M, \, \forall a\in\mathbb{R}^m,
\end{equation}
where $K, M\in\mathbb{R}^+$, and $H_\sigma$ is the Hessian of the scalar field $\sigma$, i.e., $H_\sigma(a)\in\mathbb{R}^{m\times m}$.

Now, we are ready to define the source-seeking problem.
\begin{problem}[Source-seeking]
\label{prob: ss}
	Given a signal $\sigma$ according to Definition 2, where only pointwise measurements $\sigma(p_i(t))$ are available to each robot $i \in \mathcal{V}$, and a desired precision $\epsilon \in \mathbb{R}^+$, find control actions for \eqref{eq: si} or \eqref{eq: ud} such that $\|p_c(t) - p_\sigma\| < \epsilon, \forall t \geq T$ for some finite time $T\in\mathbb{R}^+$.
\end{problem}

\vspace{0.2cm}
\section{Ascending directions for generic deployments}
\label{sec: tools}

The gradient $\nabla\sigma$ is the direction of steepest ascent of the scalar field $\sigma$, but it is not the only direction along which the field increases.

\begin{defn} \label{def: asc_dir}
	A vector $v\in\mathbb{R}^m$ is an \emph{ascending direction} at $a\in\mathbb{R}^m$ towards the source $p_\sigma$ if $\nabla\sigma(a)^\top v > 0$. A vector field $V:\mathbb{R}^m \rightarrow \mathbb{R}^m$ is an \emph{always-ascending} direction if $V(a)$ is an ascending direction at $a$ for every $a \neq p_\sigma$.
\end{defn}

The importance of this notion follows from a simple dynamical argument. Suppose the centroid evolves according to $\dot p_c = v(p_c)$, where $v(p_c)$ is an always-ascending direction. Then, along trajectories, $\frac{\mathrm{d}}{\mathrm{d}t}\sigma(p_c(t)) = \nabla\sigma(p_c)^\top v(p_c) > 0$ for all $p_c \neq p_\sigma$. Hence $\sigma(p_c(t))$ is strictly increasing until $p_c$ approaches $p_\sigma$. Under the assumption that $\sigma$ has a unique maximizer at $p_\sigma$, increasing the field value necessarily implies approaching the source. Therefore, the source-seeking Problem \ref{prob: ss} can be solved by following any always-ascending direction, not necessarily the gradient itself.

\subsection{Distributed estimation of an always-ascending direction}

Since source seeking can be achieved by following any always-ascending direction, we now construct one that can be estimated using only locally available information within the swarm. The following result, which refines and unifies \cite[Lemma 1]{acuaviva2023resilient} and \cite[Lemma 2]{acuaviva2023resilient}, formalizes this construction.

\begin{prop}[Structure of the distributed direction]
\label{prop: L_sigma}
Let $\sigma$ satisfy Definition \ref{signal}. For any non-degenerate deployment $x$ with $D = \max_{i \in \mathcal{V}}\|x_i\|$, the distributed direction
\begin{equation} \label{eq: Ls}
    L_\sigma(p_c,x) = \frac{1}{ND^2} \sum_{i=1}^N\sigma(p_c + x_i)x_i 
\end{equation}
admits the decomposition
\begin{equation} \label{eq: Ls_taylor}
	L_\sigma(p_c,x) = \underbrace{S(x)\,\nabla\sigma(p_c)}_{L_\sigma^1(p_c,x)}
	+ R(p_c,x),
\end{equation}
where $S(x) = P(x)/D^2$ is the normalized second-moment matrix of the deployment, and $L_\sigma^1(p_c,x) = S(x)\,\nabla\sigma(p_c)$ is an always-ascending direction. Moreover, the remainder satisfies
\begin{equation} \label{eq: R_bound}
    \|R(p_c,x)\| \le M D.
\end{equation}
\end{prop}

\begin{proof}
For each $i$, Taylor's theorem yields
$$
	\sigma(p_c + x_i) = \sigma(p_c) + \nabla\sigma(p_c)^\top x_i + r_i,
$$
where $|r_i|\le M\|x_i\|^2$.
Multiplying by $x_i$, summing over $i$, dividing by $ND^2$, and using $\sum_{i=1}^N x_i=0$, we obtain
$$
	L_\sigma(p_c,x) = \frac{1}{ND^2} \sum_{i=1}^N (\nabla\sigma(p_c)^\top x_i)x_i
	+ \frac{1}{ND^2} \sum_{i=1}^N r_i x_i.
$$
The first term equals $L_\sigma^1 = S(x)\nabla\sigma(p_c)$. Since the deployment is non-degenerate, $S(x)$ is positive definite; hence $\nabla\sigma(p_c)^\top L_\sigma^1 > 0$ for every $p_c \ne p_\sigma$. Thus $L_\sigma^1$ is an always-ascending direction (see \cite[Lemma 1]{acuaviva2023resilient}). For the remainder term, we have 
$$
	\left\| \frac{1}{ND^2} \sum_{i=1}^N r_i x_i \right\| 
	\le \frac{1}{ND^2} \sum_{i=1}^N M\|x_i\|^3 \le M D,
$$
which concludes the proof (see \cite[Lemma 2]{acuaviva2023resilient}).
\end{proof}

\begin{figure}[t]
	\centering
	\includegraphics[trim={0cm 0cm 0 0cm}, clip, width=1\columnwidth]{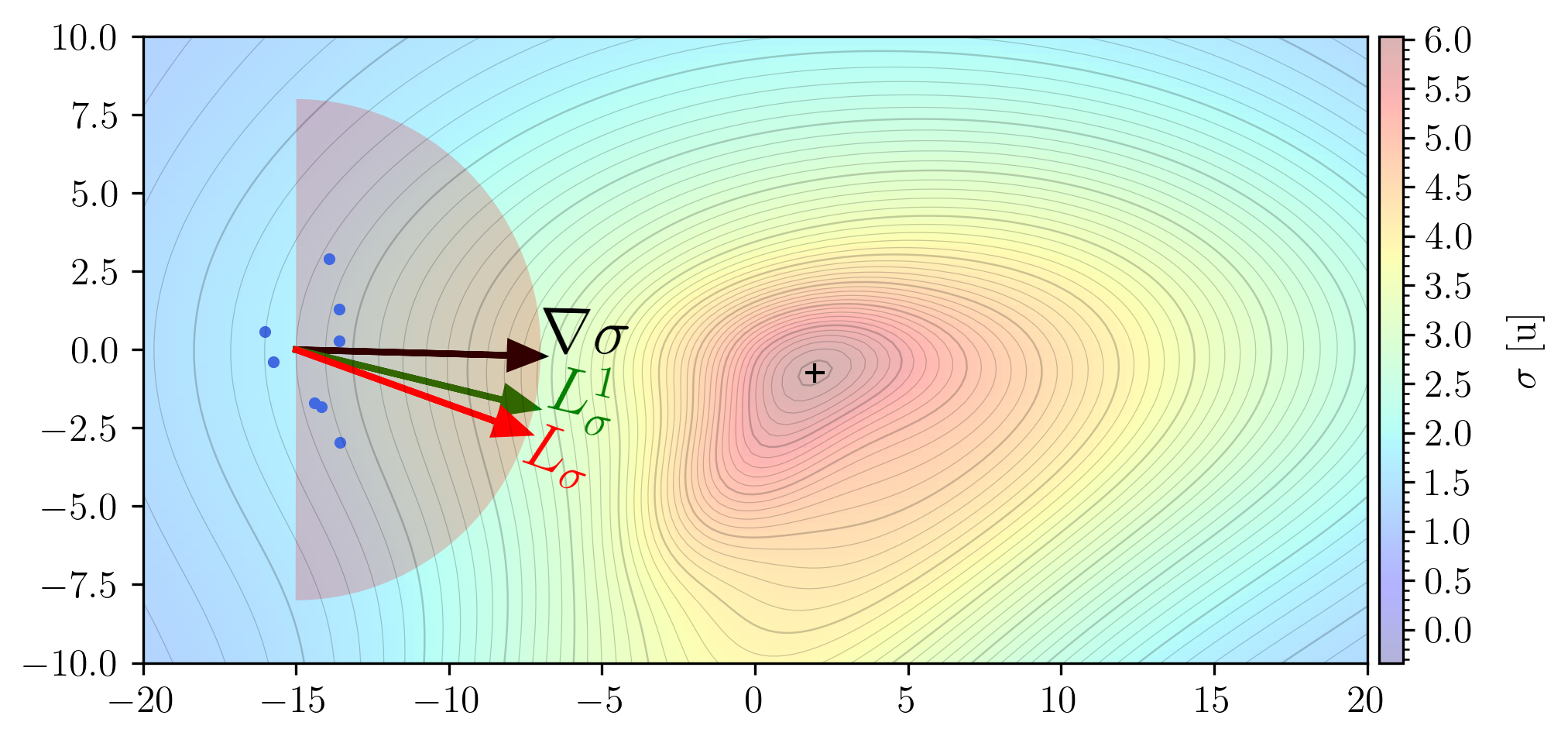}
	\caption{Illustration of the distributed direction $L_\sigma$ (red arrow) for a non-degenerate deployment of $N=8$ robots (blue dots) in a scalar field satisfying Definition \ref{signal}. The direction $L_\sigma$ approximates the always-ascending component $L_\sigma^1$ (green arrow). In general, $L_\sigma^1$ is not parallel to the gradient $\nabla\sigma$ (black arrow), but it satisfies $\nabla\sigma^\top L_\sigma^1 > 0$, meaning that it lies within the ascent half-space (red semicircle).
	}
	\label{fig: l1}
\end{figure}

\begin{figure*}
	\centering
	\includegraphics[trim={0cm 0cm 0 0cm}, clip, width=2\columnwidth]{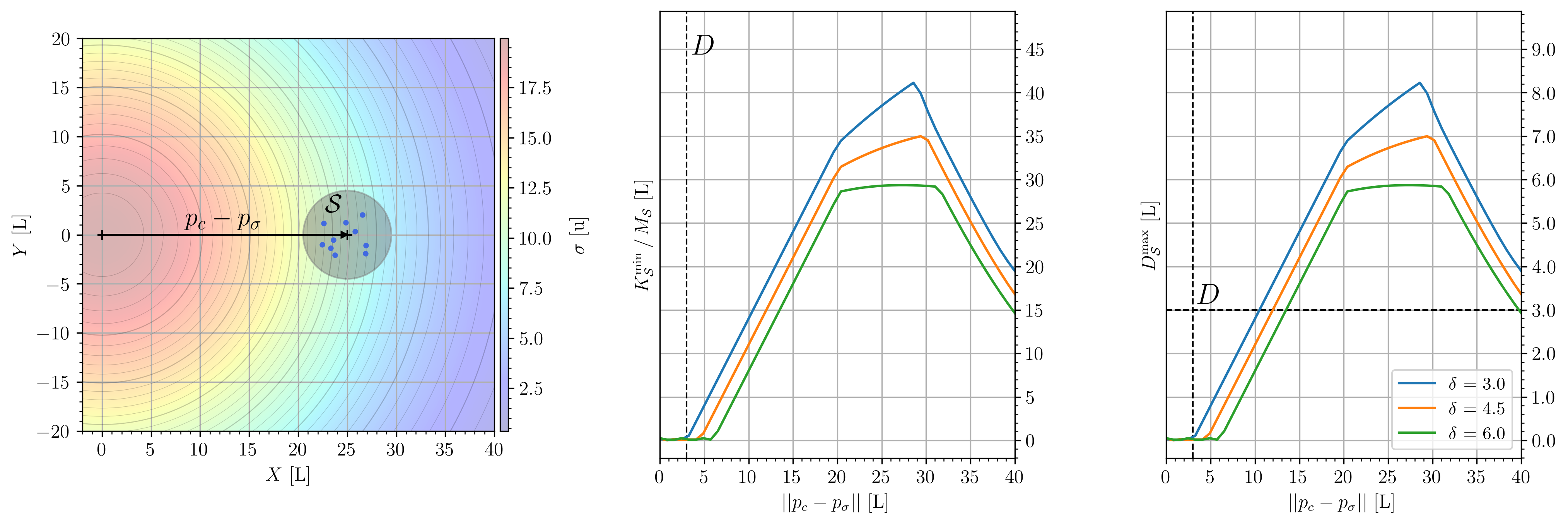}
	\caption{
	Numerical illustration of Proposition \ref{prop: D_max}. Left: Gaussian scalar field with $N=10$ robots (blue dots) in a generic deployment $x$, with $D = 3$ [L] and $\lambda_{\text{min}}(S(x)) = 0.2$. Right: evolution of the ratio $K_{\mathcal S}^{\text{min}}/M_{\mathcal S}$ and the corresponding threshold $D_{\mathcal S}^{\text{max}}$ defined in \eqref{eq: Dmax_condition}, for circular sets $\mathcal{S}$ of radius $\delta$ (grey circle) centered at $p_c$, parameterized relative to $\|p_c - p_\sigma\|$. The dashed black line indicates the fixed deployment size $D$.
	}
	\label{fig: d_max}
\end{figure*}

\begin{remark}
The distributed direction $L_\sigma$ defined in \eqref{eq: Ls} was originally introduced in \cite{brinon2015distributed} for the specific case of a circular robot formation. In that setting, the analysis exploits the symmetry of the circle to prove that $L_\sigma$ is parallel to the gradient $\nabla\sigma$. Proposition \ref{prop: L_sigma} significantly extends this result. It shows that $L_\sigma$ provides an estimate of an always-ascending direction for any non-degenerate deployment, without imposing any symmetry or particular geometric structure on the formation. 
This property relaxation provides key practical advantages, such as enabling swarm maneuvering through simple morphing of $x$ or ensuring an ascending direction even in the presence of misplaced or missing robots with respect to a reference deployment.
\end{remark}

The distributed direction $L_\sigma$ can be computed using only \emph{locally available} information, namely the centroid-relative coordinates $x_i$ and the individual measurements $\sigma(p_c + x_i)$; no global knowledge of the scalar field $\sigma$ is required. Also note that, for source seeking, the relevant feature of $L_\sigma$ is its direction rather than its magnitude (see Figure \ref{fig: l1}). Nevertheless, the normalization factor $\frac{1}{N D^2}$ ensures that $L_\sigma$, and thus the estimated always-ascending direction $L_\sigma^1$, has the same physical units as $\nabla\sigma$, thereby simplifying interpretation and highlighting the underlying structural symmetries.

\pagebreak

In general, $L_\sigma$ is not necessarily an ascending direction, since the remainder term $R$ may be large enough to yield $\nabla\sigma(p_c)^\top L_\sigma \le 0$. However, if the deployment is sufficiently compact relative to the spatial variation of the field, then $L_\sigma$ remains ascending. Indeed, $L_\sigma^1$ is always-ascending, and the perturbation term $R$ is bounded by a quantity that depends on the maximum curvature of the field and on the deployment size $D$, as stated in \eqref{eq: R_bound}.

The following result, refined from \cite[Proposition 1]{acuaviva2023resilient}, provides a sufficient condition on $D$ to ensure that $L_\sigma$ is a \emph{local} ascending direction inside a given compact set $\mathcal{S} \subset \mathbb{R}^m$ that contains the robot swarm and excludes the source $p_\sigma$.

\begin{prop} [Sufficient condition for $L_\sigma$ to be ascending]
	Let $\mathcal{S} \subset \mathbb{R}^m$ be a compact set with $p_\sigma \notin \mathcal{S}$, and assume $p_i \in \mathcal{S}$ for all $i$. Let $\sigma$ satisfy Definition \ref{signal}, and suppose that, for all $a \in \mathcal{S}$,
	$\|\nabla\sigma(a)\| \geq K_\mathcal{S}^{\text{min}}$, 
	and $\|H_\sigma(a)\| \leq 2M_\mathcal{S}$.
	If 
	\begin{equation} \label{eq: Dmax_condition}
		D < D_\mathcal{S}^{\text{max}}
		,\quad D_\mathcal{S}^{\text{max}} := \frac{K^{\text{min}}_\mathcal{S}}{M_\mathcal{S}}
		\lambda_{\text{min}}\{S(x)\},
	\end{equation}
		then $L_\sigma$ in \eqref{eq: Ls} is an ascending direction at $p_c \in \mathcal{S}$.
	\label{prop: D_max}
\end{prop}
\begin{proof}
	See \cite[Proposition 1]{acuaviva2023resilient}.
\end{proof}

\begin{example}
	\label{ex: 1}
	In Figure \ref{fig: d_max} we illustrate the result of Proposition \ref{prop: D_max} numerically. For a symmetric Gaussian signal and a generic deployment, we consider circular sets $\mathcal{S}$ centered at $p_c$ and show how the distance to the source affects the condition $D < D_\mathcal{S}^{\max}$, which guarantees that $L_\sigma$ is an ascending direction.
\end{example}

Although the signal $\sigma$ is unknown, Proposition \ref{prop: D_max} allows for the design of $D$ based on expected scenarios. For instance, similarly as in Example \ref{ex: 1}, in the case of contaminant leakage, scientists can provide the expected values of $M_\mathcal{S}$, and $K_\mathcal{S}^{\text{min}}$ within the \emph{patrolling area} $\mathcal{S}$. These values correspond to the maximum and minimum contamination thresholds at which the robot team must respond reliably. Furthermore, designing $L^1_\sigma$ to be parallel to the gradient $\nabla\sigma$ enhances the robustness of $\nabla\sigma^\top L_\sigma > 0$ with respect to $D$, as it allows for a larger deviation of $L_\sigma$ from $L^1_\sigma$, e.g., due to robots deviating from a reference deployment $x$. Thus, it is worthwhile to analyze which deployments best align $L^1_\sigma$ with the direction of $\nabla\sigma$. This is the focus of the following subsection.

\subsection{Symmetry Analysis of Generic Deployments}

The ascending direction $L^1_\sigma$ is a linear transformation of the gradient $\nabla\sigma$ through the normalized second-moment matrix $S(x)$. Therefore, the alignment between $L^1_\sigma$ and $\nabla\sigma$ depends on the eigenstructure of $S(x)$, which in turn is determined by the spatial distribution of the robots around the centroid. Indeed, if $S(x) \propto I_m$, then $L^1_\sigma$ is parallel to $\nabla\sigma$. Deployments satisfying this condition are called \emph{isotropic}, as they exhibit the same spatial spread in all directions; e.g., this is the case of regular polygons and polyhedra, as shown in \cite[Lemma 4]{acuaviva2023resilient}. Notably, this reproduces the conclusions in \cite{brinon2015distributed, brinon2019multirobot} without relying on detailed trigonometric arguments for circular or spherical formations, as regular polygons and polyhedra are naturally inscribed in circles and spheres.

Although isotropic deployments are very restrictive, they provide a useful reference to analyze the effect of morphing the swarm's spatial distribution on the direction of $L^1_\sigma$. The following result, which extends \cite[Proposition 2]{acuaviva2023resilient}, formalizes this effect.

\begin{prop}[Effect of affine deformation on $L_\sigma^1$]
\label{prop: affine_isotropic}
Let $x$ be an isotropic deployment, i.e., $P(x) = \alpha I_m$ for some $\alpha>0$. 
Consider the affine transformation $\tilde x = (I_N \otimes A)x$, with $
A \in \mathbb{R}^{m\times m}$, where $A = U\Sigma V^\top$ is the singular value decomposition of $A$. Then
\begin{equation} \label{eq: l1_svd}
	L_\sigma^1(p_c,\tilde x) 
	\propto U \Sigma^2 U^\top \nabla\sigma(p_c).
\end{equation}
\end{prop}

\begin{proof}
Since $P(x)=\alpha I_m$, the transformed second-moment matrix satisfies
$ P(\tilde x) = A P(x) A^\top = \alpha A A^\top$. Using the SVD $A = U\Sigma V^\top$, we obtain $A A^\top = U\Sigma^2 U^\top$.
Hence $P(\tilde x) = \alpha U\Sigma^2 U^\top$.
Since $L_\sigma^1 = S(\tilde x)\nabla\sigma(p_c)$ and 
$S(\tilde x) \propto P(\tilde x)$, the result follows.
\end{proof}

Proposition \ref{prop: affine_isotropic} shows that stretching an isotropic deployment reshapes the eigenvalues of its second-moment matrix while preserving orthogonality of its eigenvectors, i.e., $P(\tilde x) \propto U\Sigma^2 U^\top$. In contrast, purely orthogonal transformations $R \in \mathrm{O}(m)$ leave isotropy unchanged, since they do not modify the eigenvalues of $P(x)$.

\begin{figure}[t!]
\centering
\includegraphics[trim={0cm 0.2cm 0cm 0.7cm}, clip, width=1\columnwidth]{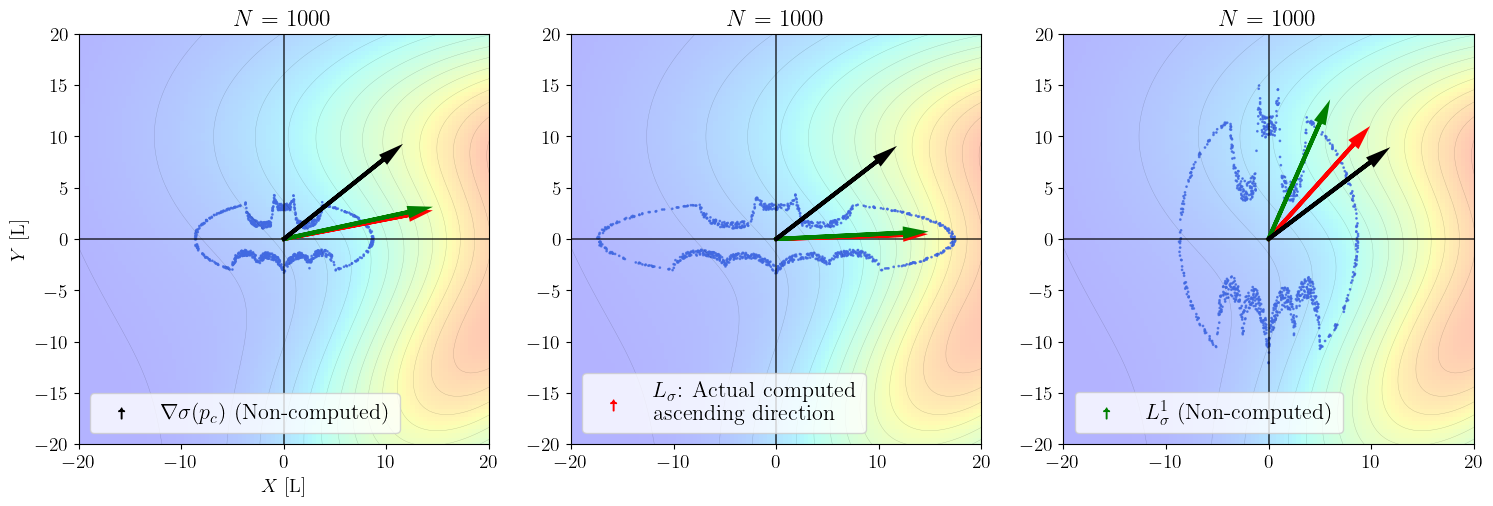}
	\caption{
		A swarm of $1000$ robots in a generic deployment (left) is stretched horizontally (center) and vertically (right) to illustrate how the direction of $L_\sigma$ varies with the shape of the swarm. In all three figures, the black arrows represent the gradient $\nabla\sigma$, the green arrows correspond to $L^1_\sigma$, and the red arrows denote the direction $L_\sigma$ computed by the robots. All arrows are normalized for visualization purposes, as only their directions are relevant.
	}
\label{fig: Batman}
\end{figure}

The matrix $U\Sigma^2 U^\top$ is the spectral decomposition of a positive definite matrix whose eigenvectors (columns of $U$) define the principal \emph{axes of stretching}, and whose eigenvalues ($\Sigma^2$) quantify the variance along those axes. Consequently, from \eqref{eq: l1_svd}, the components of the gradient along stretched directions are amplified in $L_\sigma^1$, whereas those along compressed directions are attenuated, as shown in Figure \ref{fig: Batman}.
This provides a geometric mechanism to bias the ascent direction: by anisotropically scaling an otherwise isotropic formation, one can steer $L_\sigma^1$ toward desired directions, thereby influencing the collective motion of the swarm when it follows $L_\sigma$.

The eigenvalues of $S(x)$ contain even more information. The minimum eigenvalue $\lambda_{\text{min}}(S(x))$ quantifies the least directional spread of the deployment and plays a central role in the ascent condition. Indeed, $\lambda_{\text{min}}(S(x)) = 0$ if and only if the deployment is degenerate, in which case $L_\sigma^1$ is not an ascending direction. A larger values of $\lambda_{\text{min}}(S(x))$ enhance the robustness of $L_\sigma$ with respect to the remainder term $R$, as it allows for a larger deviation of $L_\sigma$ from $L_\sigma^1$ while still ensuring $\nabla\sigma^\top L_\sigma > 0$, as shown in Proposition \ref{prop: D_max}.

It should be noted that $\lambda_{\text{min}}(S(x))$ is not necessarily increasing with the number of robots $N$. For instance, if all robots are aligned along a line, then $\lambda_{\text{min}}(S(x)) = 0$ regardless of $N$. However, for sufficiently uniform deployments, increasing $N$ can improve the \emph{robustness} of the ascent direction. In particular, the following result shows that the variation of $\lambda_{\text{min}}(S(x))$ induced by the loss of a single robot scales as $O(1/N)$.

\begin{prop}[Robustness to robot removal]
\label{prop: single_failure_weyl}
Let $S(x) = P(x)/D^2$ be the normalized second-moment matrix of a deployment $x$ with $N \ge 2$ robots, and let $x^{(-j)} \in \mathbb{R}^{m(N-1)}$ denote the deployment after removing robot $j$, with barycentric coordinates recomputed with respect to the new centroid. Then
$$
\big|
\lambda_{\min}(S(x)) 
-
\lambda_{\min}(S(x^{(-j)}))
\big|
\le
\frac{4}{N-1}.
$$
\end{prop}
\begin{proof}
See Subsection \ref{proof: single_failure_weyl} in the Appendix.
\end{proof}

Exploiting the symmetry of the deployment not only provides insights into the first-order direction $L_\sigma^1$, but also yields structural simplifications for the higher-order terms in the Taylor expansion of $L_\sigma$. In particular, the following result shows that centrally symmetric deployments cause all even-order contributions to $L_\sigma$ to vanish. This strengthens the alignment between $L_\sigma$ and $L_\sigma^1$ and reduces the impact of the curvature term in the field, effectively improving the accuracy of the swarm's ascent direction.

\begin{prop}[Cancellation of even-order terms under central symmetry] \label{prop: cancellation}
	Let $x$ be a centrally symmetric deployment, i.e.
	for every $i$ there exists $j$ such that $x_j = -x_i$. Let $\sigma \in C^\infty(\mathbb{R}^m)$. Then the Taylor expansion of $L_\sigma$ around $p_c$ contains only odd-order derivatives of $\sigma$. More precisely,
	$$
	L_\sigma(p_c,x)
	=
	\frac{1}{ND^2}
	\sum_{\substack{k\ge 0 \\ k \text{ odd}}}
	\frac{1}{k!}
	\sum_{i=1}^N
	D^k \sigma(p_c)[x_i^{\otimes k}]\, x_i,
	$$
	where $D^k\sigma(p_c)$ denotes the $k$-th Fréchet derivative and
	$x_i^{\otimes k}$ the $k$-fold tensor product of $x_i$.
\end{prop}

\begin{proof}
Using the Taylor expansion of $\sigma$ at $p_c$, for each $i$,
$$
\sigma(p_c + x_i)
=
\sum_{k=0}^{\infty}
\frac{1}{k!}
D^k \sigma(p_c)[x_i^{\otimes k}].
$$
Multiplying by $x_i$ and summing over $i$ yields
$$
L_\sigma(p_c,x)
=
\frac{1}{ND^2}
\sum_{k=0}^{\infty}
\frac{1}{k!}
\sum_{i=1}^N
D^k \sigma(p_c)[x_i^{\otimes k}]\, x_i.
$$

Since the deployment is centrally symmetric, for every $x_i$
there exists $-x_i$. Hence terms can be grouped in pairs:
$$
D^k \sigma(p_c)[x^{\otimes k}]\, x
+
D^k \sigma(p_c)[(-x)^{\otimes k}]\, (-x).
$$

Because $(-x)^{\otimes k} = (-1)^k x^{\otimes k}$, we obtain
$$
D^k \sigma(p_c)[(-x)^{\otimes k}]\, (-x)
=
(-1)^{k+1}
D^k \sigma(p_c)[x^{\otimes k}]\, x.
$$
Thus the pair contributes $\left(1 + (-1)^{k+1}\right)
D^k \sigma(p_c)[x^{\otimes k}]\, x$. If $k$ is even, then $1 + (-1)^{k+1} = 0$, so the term vanishes.
If $k$ is odd, the contribution doubles. Therefore all even-order Taylor terms cancel, and only odd-order derivatives remain.
\end{proof}

\begin{coroll} \label{col: quad}
	If the deployment is centrally symmetric and $\sigma$ is a quadratic scalar field, then it directly follows from Proposition \ref{prop: cancellation} that $L_\sigma(p_c,x) = L_\sigma^1(p_c,x)$.
\end{coroll}

Although simple, quadratic scalar fields arise naturally in many physical contexts, such as the logarithm of Gaussian fields or potentials following $\frac{1}{r^2}$ laws. In these cases, the remainder terms vanish entirely for centrally symmetric deployments, so the distributed direction $L_\sigma$ coincides exactly with the always-ascending direction $L_\sigma^1$ (Corollary \ref{col: quad}). Moreover, for centrally symmetric deployments, $S(x) \propto I_m$, which implies that $L_\sigma^1$ is parallel to the gradient $\nabla\sigma$.



\pagebreak

\section{Distributed architecture of the proposed source-seeking system}
\label{sec: arch}

Up to this point, the computation of $L_\sigma$ in \eqref{eq: Ls} has assumed a central robot collecting all terms $\sigma(p_c + x_i)x_i$, making the swarm dependent on a single point of failure. While one could assign backup computing robots, a more resilient approach is to perform all necessary computations in a fully distributed manner. 

The results presented in this section are fully novel with respect to our previous conference work \cite{acuaviva2023resilient}, which relied on central computation and complete-graph communication. Here, we develop a leaderless, distributed architecture where robots can join or leave the swarm at any time without compromising the mission. In order to provide a self-contained and distributed source-seeking solution, we show the following items in the upcoming subsections. Firstly, the compatibility of the source-seeking solution for single-integrator robots with a displacement-based formation controller. Secondly, the distributed estimation of the swarm centroid, i.e., each robot estimates its corresponding $x_i$ in (\ref{eq: Ls}). This estimator is based on the \emph{dual} version of the above formation controller. Lastly, the distributed calculation of the ascending direction (\ref{eq: Ls}).

We illustrate the closed-loop system architecture in Figure \ref{fig: archi}, where for single-integrator robots, the ascending-direction tracking occurs \emph{instantaneously} We will see that the exponential convergence of each algorithm facilitates a slow-fast system interconnection, with centroid estimation being the fastest, and tracking being the slowest, since the constant speed for the robots can be set arbitrarily. The tracking controller for the 2D unicycle will be presented in the upcoming Section \ref{sec: uni}.

\subsection{Compatibility with a displacement-based formation control}
\label{sub: fc}

We briefly establish that the source-seeking control law is compatible with standard displacement-based formation control for single-integrator robots. This compatibility is important for two reasons: (i) it allows the swarm to maintain a \emph{desired shape} while seeking the source to assist on the calculation of $L_\sigma$, and (ii) it motivates the distributed estimators introduced in subsequent subsections through a \emph{duality} principle.

Consider the displacement-based formation controller for single-integrator robots \eqref{eq: si}:
\begin{equation}
u_f^i = -k_f \sum_{j \in \mathcal{N}_i} \Big[(p_i - p_j) - (p_i^* - p_j^*)\Big], \quad \forall i \in \mathcal{V},
\label{eq: uf}
\end{equation}
where $k_f > 0$ and $(p_i^* - p_j^*)$ are the desired relative positions. In compact form, $u_f = -k_f \bar{L}(p - p^*)$, where $L$ is the Laplacian matrix of $\mathcal{G}$. Since the graph $\mathcal{G}$ is assumed connected and undirected, this controller drives the formation to the desired shape exponentially fast \cite{oh2015survey}, though the absolute position converges to some translate of $p^*$ since $\mathbf{1}_N \in \ker(L)$. In fact, if the desired shape $p^*$ is written with respect to $p_c$, then $\lim_{t\to\infty} x(t) = p^*$ exponentially fast \cite{de2020maneuvering}.

The key observation is that the source-seeking velocity $\frac{L_\sigma}{\|L_\sigma\|}$ lies in $\ker(\overline{L})$ when applied uniformly to all robots, i.e.,
$$
\overline{L}\left(\mathbf{1}_N \otimes \frac{L_\sigma}{\|L_\sigma\|}\right) = 0.
$$
Therefore, the combined control law
\begin{equation}
u = \left(\mathbf{1}_N \otimes \frac{L_\sigma}{\|L_\sigma\|} \right) + u_f,
\label{eq: u}
\end{equation}
decouples the source-seeking motion (centroid translation) from formation stabilization (shape control). The formation controller $u_f$ preserves the centroid since $\sum_{i=1}^N u_f^i = 0$, while the source-seeking term translates the entire formation along the ascending direction. Indeed, since the convergence of \eqref{eq: uf} is exponentical, $u_f$ can be seen as a perturbation that vanishes exponentially fast, and the analysis of \eqref{eq: u} to solve Problem \ref{prob: ss} is similar to the one in \cite[Theorem 1]{acuaviva2023resilient}.

\begin{figure}
\centering
\resizebox{0.48\textwidth}{!}{
    \begin{tikzpicture}[
        box/.style = {draw, rectangle, minimum width=3cm, minimum height=1.5cm, text centered, rounded corners},
        arrow/.style = {draw, -{Latex[length=2mm, width=2mm]}}
        ]

		\node[block_red, text width=3cm] (f1) {Centroid estimation protocol \eqref{eq: esti}};
        \node[block_red, text width=3.5cm, right=1cm of f1] (f2) {Ascending direction estimation protocol \eqref{eq: mu}};
		\node[block,right=3cm of f2, text width=3cm] (tracking) {Ascending direction tracking + formation control \eqref{eq: u} for s.i. or \eqref{eq_omegai} for unicycles};
		\node[block_kin,below=1cm of tracking, text width=3cm] (kin) {Robot dynamics \eqref{eq: si} or \eqref{eq: ud}};
		\node[block_nei, left=1.0cm of kin] (nei) {Neighbors' closed loop};

        \draw[arrow] (f1) -- node[above]{$\hat x_i$} (f2);

		\draw[arrow] (f2) -- node[above]{$\frac{\mu}{\|\mu\|} = \frac{L_\sigma}{\|L_\sigma\|}$} (tracking);
		\draw[arrow] (tracking) -- node[pos=0.15, below, xshift=0.5cm]{$u_i, \omega_i$} (kin) ;
    	\draw[arrow] (kin) -- (nei);
		\draw[arrow] (nei) -| node[pos=0.9, below, xshift=1cm]{$\hat x_j, (p_i - p_j)$} (f1.south);
		\draw[arrow] (nei) -| node[pos=0.9, below, xshift=0.6cm]{$\hat\mu_j, \mu_j$} (f2.south);
		\draw [arrow] (kin.east) -| ++(0.5,3.8) -| node [pos=0.5, below, xshift=0.7cm] {$\sigma(p_i)$} (f2.north);
		\draw[arrow] (nei) |- node[pos=0.65, below, xshift=0.3cm]{$(p_i - p_j)$} ([yshift=-0.4cm]tracking.west);
    \end{tikzpicture}
	}
	\caption{
		Closed-loop system of an individual robot showing all the algorithms and inputs/outputs involved in the proposed source-seeking solution. The exponential convergence of the three blocks on top facilitates their interconnection in a slow-fast fashion. All (neighboring) robots run the same closed-loop system represented as a dashed block, and note how their (relative and absolute) positions are also in closed-loop; this is represented by the arrow going from "Robot dynamics" to "Neighbors' close loop" block.}
\label{fig: archi}
\end{figure}

\subsection{Distributed estimation of the centroid}

We now present the algorithm that estimates the centroid-relative coordinates $x_i$ in a distributed manner. The proposed estimator is based on well-established consensus dynamics from the formation control and distributed estimation literature \cite{yang2018, de2020maneuvering, bullo2020lectures}. In particular, the following lemma summarizes standard convergence results for consensus-based estimators that will be instrumental for the singular perturbation analysis in Theorem~\ref{th: dis}.

\begin{lemma}[Consensus-based relative estimation]
\label{lem: consensus_estimation}
Consider a connected undirected graph $\mathcal{G} = (\mathcal{V}, \mathcal{E})$ and let $y_i \in \mathbb{R}^m$ be quantities associated with each agent $i \in \mathcal{V}$. Define $y_c := \frac{1}{N}\sum_{i=1}^N y_i$ as their centroid. The distributed estimator
\begin{equation}
\dot{\hat{y}}_i = -\sum_{j \in \mathcal{N}_i} \Big[(\hat{y}_i - \hat{y}_j) - (y_i - y_j)\Big], \quad \forall i \in \mathcal{V},
\label{eq: general_estimator}
\end{equation}
satisfies:
\begin{enumerate}[label=\roman*)]
    \item The centroid $\hat{y}_c(t) := \frac{1}{N}\sum_{i=1}^N \hat{y}_i(t)$ is invariant, i.e., $\dot{\hat{y}}_c(t) = 0$.
    \item The estimator converges exponentially to $\hat{y}_i = y_i + c$, where $c = \hat{y}_c(0) - y_c$ is determined by the initial conditions.
    \item If $\hat{y}_c(0) = 0$, then $\hat{y}_i(t)$ converges exponentially fast to $y_i - y_c$.
\end{enumerate}
\end{lemma}
\begin{proof}
Writing the compact form $\dot{\hat{y}} = -\bar{L}\hat{y} + \bar{L}y$, we have $\mathbf{1}_N^\top \dot{\hat{y}} = 0$ since $\mathbf{1}_N^\top L = 0$, proving (i). The equilibrium condition $\bar{L}(\hat{y} - y) = 0$ implies $\hat{y}_i - y_i = c$ for some $c \in \mathbb{R}^m$ common to all agents, i.e., $\hat y = y + \mathbf{1}_N \otimes c$. Taking the average over all agents gives $\hat{y}_c = \frac{1}{N}\sum_{i=1}^N \hat y_i = y_c + c$. Since $\dot{\hat{y}}_c(t) = \dot{\hat{y}}_c(0)$ by the invariance in (i), at equilibrium we have $\hat{y}_c(0) = y_c + c$. Solving for $c$ yields $c = \hat{y}_c(0) - y_c$, proving (ii). Setting $\hat{y}_c(0) = 0$ gives $c = -y_c$, hence $\hat{y}_i \to y_i - y_c$, proving (iii). Exponential convergence follows from the spectral properties of the graph Laplacian \cite{yang2018, bullo2020lectures,de2020maneuvering}: for connected graphs, $\lambda_2(L) > 0$, and the error $\tilde{y} = \hat{y} - (y + \mathbf{1}_N \otimes c)$ satisfies $\dot{\tilde{y}} = -\bar{L}\tilde{y}$, which converges to zero exponentially with rate $\lambda_2(L)$ in the subspace orthogonal to $\ker(L)$.
\end{proof}

We now apply Lemma \ref{lem: consensus_estimation} to estimate the centroid-relative coordinates. Recall from Section \ref{sub: fc}, the formation control problem has its dual in the relative localization problem \cite{oh2015survey, zhang2011optimal}. In the formation controller \eqref{eq: uf}, the \emph{hardware} term associated with the robot measurements $(p_i - p_j)$ is dynamic, while the \emph{software} term $(p_i^* - p_j^*)$ defining the desired shape is fixed; in the estimator, these roles are reversed. The centroid estimator is given by
\begin{equation}
\dot{\hat{x}}_i(t) = -\sum_{j\in\mathcal{N}_i} \Big[(\hat{x}_i(t) - \hat{x}_j(t)) - (x_i - x_j)\Big], \quad \forall i\in\mathcal{V},
\label{eq: esti}
\end{equation}
where the hardware measurements $(x_i - x_j) = (p_i - p_j)$ serve as fixed references (or slowly-varying inputs in practice), and the software estimates $\hat{x}_i$ are communicated between neighboring robots, i.e., between every pair $(i,j)\in\mathcal{E}$.

\begin{coroll}[Distributed centroid estimation]
\label{cor: centroid}
Let $y_i = x_i$ in Lemma \ref{lem: consensus_estimation}, where $x_i$ are the centroid-relative coordinates satisfying $x_c = 0$ by definition. If $\hat{x}_c(0) = 0$ (e.g., by setting $\hat{x}_i(0) = 0$ for all $i$), then $\hat{x}_i(t)$ converges exponentially fast to $x_i$.
\end{coroll}

\begin{remark}[Time-varying measurements] \label{rem: time_varying}
In practice, the relative positions $(x_i - x_j)$ evolve as the robots move. The estimator \eqref{eq: esti} can track such time-varying signals provided the robots move slowly relative to the estimator dynamics. This time-scale separation is formalized in Section \ref{sec: sf}. More sophisticated tracking techniques from \cite{kia2019tutorial} can further improve performance.
\end{remark}

The result of Lemma \ref{lem: consensus_estimation} can also be used to estimate the relative positions of two non-neighboring robots, or more generally, to localize all the robots with respect to a chosen one. For example, once robot $k$ estimates its centroid-relative coordinate $x_k$, we can initialize a new estimation network by setting $\sum_{i=1}^N \hat x_i(0) = -x_k$. In particular, robot $k$ assigns $\hat x_k(0) = -x_k$, while the rest of the robots set their initial conditions to zero; thus $\lim_{t\to\infty}\hat x(t) = x - (\mathbf{1}_N \otimes x_k)$. We summarize this derivation with the following corollary. 

\begin{coroll}[Distributed localization with respect to a reference robot]
\label{cor: localization}
Let $y_i = x_i$ in Lemma \ref{lem: consensus_estimation} and choose a reference robot $k \in \mathcal{V}$. If $\hat{x}_c(0) = -x_k$ (e.g., by setting $\hat{x}_k(0) = -x_k$ and $\hat{x}_i(0) = 0$ for all $i \neq k$), then $\hat{x}_i(t)$ converges exponentially fast to $x_i - x_k = p_i - p_k$.
\end{coroll}

\subsection{Distributed calculation of the ascending direction}
\label{sec: mu}

As we have seen in (\ref{eq: u}), so far for single-integrator robots, the swarm can track the direction of $L_\sigma$. In this subsection, we are going to show that the direction of the vector $L_\sigma$ as in (\ref{eq: Ls}) can be estimated with a consensus algorithm distributively. We first note that $N$ and $D^2$ are not necessary to know the direction of $L_\sigma$; therefore if we define
\begin{equation}
\label{eq: mut}
	\mu_c(t) := \frac{1}{N}\sum_{i=1}^N \mu_i(t) = \frac{1}{N}\sum_{i=1}^N \sigma(p_c(t) + x_i(t))x_i(t),
\end{equation}
then we have that $\frac{L_\sigma}{\|L_\sigma\|} = \frac{\mu_c}{\|\mu_c\|}$, i.e., the signals $\mu_c(t)$ and $L_\sigma(t)$ point at the same direction. Hence, if we set $\mu_i(0) = \sigma(p_c(0) + x_i(0))x_i(0)$, we have that $\lim_{t\to\infty}\mu_i(t) \to \mu_c(0)$ exponentially fast if a connected network within the robot swarm follows the consensus protocol
\begin{equation}
	\label{eq: mu}
	\begin{cases}
		\dot\mu_i(t) &= \sum_{j\in\mathcal{N}_i} \left(\mu_j(t) - \mu_i(t)\right) \\ \mu_i(0) &= \sigma(p_c(0) + x_i(0))x_i(0)
	\end{cases}, \, \forall i\in \mathcal{V}.
\end{equation}
However, the protocol (\ref{eq: mu}) cannot run without resets each time suitable values of $\sigma(p_i)$ and $\hat x_i$ are available for the whole network. Moreover, the resulting ascending direction to feed the tracking controller has to be refreshed after an almost asymptotic value of $\mu_i(t) \to \mu_c(0)$ has been reached.

In order to have a dynamic protocol for the estimation of $\mu_c(t)$ that does not need to be reset, we leverage Lemma \ref{lem: consensus_estimation}. Each robot updates a software variable $\hat{\mu}_i(t)$ using the locally computed values $\mu_i(t)$ and those received from neighbors, following the estimation protocol in \eqref{eq: general_estimator}:
\begin{equation}
\label{eq: estimu}
	\scalemath{0.9}{
	\begin{cases}
		\dot{\hat \mu}_i(t) &= -\sum_{j\in\mathcal{N}_i} \Big((\hat \mu_i(t) - {\hat \mu}_j(t)) - (\mu_i - \mu_j) \Big) \\
		\hat\mu_i(0) &= 0, \qquad \forall i\in\mathcal{V}.
	\end{cases}
	}
\end{equation}
Note that each robot $i$ must receive $\mu_j$ and $\hat{\mu}_j$ from every neighbor $j \in \mathcal{N}_i$. 

\begin{coroll}[Distributed ascending direction estimation]
\label{cor: ascending}
Assume all robots are stationary, i.e., $\dot{p} = 0$, so that $\mu_i = \sigma(p_i)x_i$ is constant for all $i \in \mathcal{V}$. Let $y_i = \mu_i$ in Lemma \ref{lem: consensus_estimation}, with centroid $y_c = \mu_c = \frac{1}{N}\sum_{i=1}^N \mu_i$. If $\hat{\mu}_c(0) = 0$ (e.g., by setting $\hat{\mu}_i(0) = 0$ for all $i$), then $\hat{\mu}_i(t)$ converges exponentially fast to $\mu_i - \mu_c$. Consequently, each robot can compute $\mu_c^i := \mu_i - \hat{\mu}_i \to \mu_c$, providing a distributed estimate of the direction $L_\sigma / \|L_\sigma\| = \mu_c / \|\mu_c\|$.
\end{coroll}

\begin{remark}[Communication and computational complexity]
	The per-robot computational cost of both estimators \eqref{eq: esti} and \eqref{eq: estimu} is $O(|\mathcal{N}_i|)$, scaling linearly with the number of neighbors rather than the swarm size. Communication requirements are similarly local: each robot exchanges $\hat{x}_i$ and $\hat{\mu}_i$ with its neighbors, yielding $O(|\mathcal{E}|)$ total messages per iteration. For sparse graphs with bounded degree, both complexities are $O(1)$ per robot, ensuring scalability to large swarms. However, the algebraic connectivity $\lambda_2(L) \in \mathbb{R}$ determines the convergence rate; sparser graphs require higher update frequencies to achieve a fixed convergence time, increasing the overall communication and computational load.
\end{remark}

As with the centroid estimator (see Remark \ref{rem: time_varying}), the stationarity assumption $\dot{p} = 0$ is relaxed in the following subsection through a time-scale separation argument.

\pagebreak

\subsection{Slow-fast closed-loop system}
\label{sec: sf}

To relax the stationarity assumption on $x_i(t)$ and $\mu_i(t)$, we exploit the exponential convergence of both estimators established in Lemma \ref{lem: consensus_estimation}. While more sophisticated dynamic tracking techniques from \cite{kia2019tutorial} could be incorporated, we adopt the following time-scale separation assumptions for clarity and conciseness:
\begin{itemize}
	\item The robots move slowly relative to the estimator dynamics, so that the signals $\sigma(p_i)$ and the relative positions $(p_i - p_j)$ evolve on a slower time scale.
	\item The centroid estimator \eqref{eq: esti} operates faster than the ascending direction estimator \eqref{eq: estimu}, which in turn is faster than the robot motion.
\end{itemize}

Under these assumptions, we interconnect \eqref{eq: esti}, \eqref{eq: estimu}, and \eqref{eq: u}, yielding the following slow-fast closed-loop system (see Figure \ref{fig: archi}) for single-integrator robots:

\begin{equation} \label{eq: sf}
\left\{
    \begin{aligned}
        \epsilon_x \dot{\hat{x}}_i &= 
        -\sum_{j \in \mathcal{N}_i}\Big[(\hat{x}_i - \hat{x}_j) - (x_i - x_j)\Big],\\[1mm]
        \epsilon_\mu \dot{\hat{\mu}}_i &= 
        -\sum_{j \in \mathcal{N}_i}\Big[(\hat{\mu}_i - \hat{\mu}_j) - (\mu_i - \mu_j)\Big],\\[1mm]
        \dot{p}_i &= 
        \frac{\mu^i_c}{\|\mu^i_c\|}
        - k_f\sum_{j \in \mathcal{N}_i}\Big[(p_i - p_j) - (p^i_d - x^j_d)\Big],\\[1mm]
		\mu_i &= \sigma(p_i)\hat x_i, \quad \mu_c^i = \mu_i - \hat \mu_i \\[1mm]
		\hat x(0) &=0, \quad \hat \mu(0)=0,
    \end{aligned}
\right.
\end{equation}
where $\epsilon_x, \epsilon_\mu > 0$ are two sufficiently
small constants to tune the slow-fast system. The available measurements are the scalar field readings $\sigma(p_i(t))$ and the relative positions $(p_i(t)-p_j(t))$, $(i,j)\in\mathcal{E}$. The desired formation is defined by $x_d = p_d - p_c \in \mathbb{R}^{mN}$. The following theorem establishes that this closed-loop system solves Problem~\ref{prob: ss}.


\begin{theorem}
\label{th: dis}
	Consider the signal $\sigma$ as defined in Definition \ref{signal} and a swarm of $N$ robots with dynamics (\ref{eq: si}) whose sensing and communication topology is given by an undirected and connected graph $\mathcal{G}$. Assume that the initial deployment $x(0)$ is non-degenerate and that the conditions of Proposition \ref{prop: D_max} hold for an annulus shell $\mathcal{S}$ around $p_\sigma$. If all robots remain within $\mathcal{S}$ during the mission, then there exist $\epsilon_x^* > 0$ and $\epsilon_\mu^* > 0$ such that for all $0 < \epsilon_x < \epsilon_x^*$ and $0 < \epsilon_\mu < \epsilon_\mu^*$ with $\epsilon_x < \epsilon_\mu$, the closed-loop system (\ref{eq: sf}) solves Problem \ref{prob: ss}.
\end{theorem}
\begin{proof}
	To analyze the fastest subsystem, introduce the fast time scale
	$\tau = t/\epsilon_x$.  
	Then system~\eqref{eq: sf} becomes
\begin{equation*}
\scalemath{0.9}{
\left\{
    \begin{aligned}
        \frac{\mathrm{d}\hat{x}_i}{\mathrm{d}\tau} &= 
        -\sum_{j \in \mathcal{N}_i}\Big[(\hat{x}_i - \hat{x}_j) - (x_i - x_j)\Big],\\[1mm]
        \frac{\mathrm{d}\hat{\mu}_i}{\mathrm{d}\tau} &= 
        -\frac{\epsilon_x}{\epsilon_\mu}
        \sum_{j \in \mathcal{N}_i}\Big[(\hat{\mu}_i - \hat{\mu}_j) - (\mu_i - \mu_j)\Big],\\[1mm]
        \frac{\mathrm{d}p_i}{\mathrm{d}\tau} &= 
        \epsilon_x\left(
        \frac{\mu^i_c}{\|\mu^i_c\|}
        - k_f\sum_{j \in \mathcal{N}_i}\Big[(p_i - p_j) - (p^i_d - x^j_d)\Big]\right).
    \end{aligned}
\right.
}
\end{equation*}

Setting $\epsilon_x = 0$ yields the boundary-layer system
\begin{equation}\label{eq:1_layer}
\frac{\mathrm{d}\hat{x}_i}{\mathrm{d}\tau} = 
-\sum_{j \in \mathcal{N}_i}\Big[(\hat{x}_i - \hat{x}_j) - (x_i - x_j)\Big],
\end{equation}
with $(\hat{\mu}, p)$ frozen. By Lemma~\ref{lem: consensus_estimation} and Corollary~\ref{cor: centroid}, this system has an isolated exponentially stable equilibrium at $\hat{x}_i = x_i$ (using $\hat{x}_i(0) = 0$ and $x_c = 0$). Substituting this equilibrium into the remaining dynamics yields the reduced system
\begin{equation}\label{eq:reduced_1}
\left\{
    \begin{aligned}
        \epsilon_\mu \dot{\hat{\mu}}_i &= 
        -\sum_{j \in \mathcal{N}_i}\Big[(\hat{\mu}_i - \hat{\mu}_j) - (\mu_i - \mu_j)\Big],\\[1mm]
        \dot{p}_i &= 
        \frac{\mu^i_c}{\|\mu^i_c\|}
        - k_f\sum_{j \in \mathcal{N}_i}\Big[(p_i - p_j) - (p^i_d - x^j_d)\Big],
    \end{aligned}
\right.
\end{equation}
where $\mu_i = \sigma(p_i)x_i$ and $x_i = p_i - p_c$. By \cite[Theorem 11.4]{khalil2015nonlinear}, there exists $\epsilon_x^* > 0$ such that for all $0 < \epsilon_x < \epsilon_x^*$, the stability properties of the full system are determined by those of the reduced system \eqref{eq:reduced_1}.

System \eqref{eq:reduced_1} is itself singularly perturbed with small parameter $\epsilon_\mu$. Introduce the time scale $\tilde{\tau} = t/\epsilon_\mu$ and set $\epsilon_\mu = 0$ to obtain the boundary-layer system
\begin{equation}\label{eq:2_layer}
\frac{\mathrm{d}\hat{\mu}_i}{\mathrm{d}\tilde{\tau}} = 
-\sum_{j \in \mathcal{N}_i}\Big[(\hat{\mu}_i - \hat{\mu}_j) - (\mu_i - \mu_j)\Big],
\end{equation}
with $p$ frozen. By Lemma~\ref{lem: consensus_estimation} and Corollary~\ref{cor: ascending}, this system has an isolated exponentially stable equilibrium at $\hat{\mu}_i = \mu_i - \mu_c$ (using $\hat{\mu}_i(0) = 0$). Substituting this equilibrium yields the second reduced system
\begin{equation}\label{eq:reduced_2}
\dot{p}_i = 
\frac{\mu_c}{\|\mu_c\|}
- k_f\sum_{j \in \mathcal{N}_i}\Big[(p_i - p_j) - (p^i_d - x^j_d)\Big].
\end{equation}
By \cite[Theorem 11.4]{khalil2015nonlinear}, there exists $\epsilon_\mu^* > 0$ such that for all $0 < \epsilon_\mu < \epsilon_\mu^*$, the stability of \eqref{eq:reduced_1} is determined by that of \eqref{eq:reduced_2}.

The formation controller in \eqref{eq:reduced_2} is exponentially convergent: the term $-k_f\sum_{j \in \mathcal{N}_i}[(p_i - p_j) - (p^i_d - x^j_d)]$ drives the deployment $x$ to the desired shape $x_d$ exponentially fast, while the term $\frac{\mu_c}{\|\mu_c\|}$ provides a common velocity to all robots that does not affect the relative positions (since $\bar{L}(\mathbf{1}_N \otimes v) = 0$ for any $v \in \mathbb{R}^m$).

Combining the above, for all $0 < \epsilon_x < \epsilon_x^*$ and $0 < \epsilon_\mu < \epsilon_\mu^*$ with $\epsilon_x < \epsilon_\mu$, the closed-loop system \eqref{eq: sf} is exponentially stable. In particular, $\dot{p}_i$ converges exponentially to $\frac{\mu_c}{\|\mu_c\|}$ for all $i \in \mathcal{V}$. By the conditions of Proposition~\ref{prop: D_max}, $\frac{\mu_c}{\|\mu_c\|}$ is an ascending direction whenever the robots are inside $\mathcal{S}$. Since the robots remain within $\mathcal{S}$ by assumption, the centroid velocity $\dot{p}_c = \frac{1}{N}\sum_{i=1}^N \dot{p}_i$ converges to an ascending direction, ensuring that $\sigma(p_c(t))$ strictly increases until $p_c$ enters the $\epsilon$-neighborhood of $p_\sigma$. This solves Problem~\ref{prob: ss}.
\end{proof}

\begin{remark}[Role of estimator initialization]
\label{rem: initialization}
The initializations $\hat{x}_i(0) = 0$ and $\hat{\mu}_i(0) = 0$ are essential for the stability analysis. As stated in Lemma \ref{lem: consensus_estimation}, the consensus-based estimators \eqref{eq: esti} and \eqref{eq: estimu} naturally admit invariant \emph{manifolds of equilibria} parameterized by the initial centroid of the estimates. By choosing $\hat{x}_i(0) = 0$ for all $i$, together with $x_c = 0$ by definition, these manifolds collapse to the \emph{isolated equilibria} $\hat{x}_i = x_i$ and $\hat{\mu}_i = \mu_i - \mu_c$, respectively. This reduction is necessary because \cite[Theorem~11.4]{khalil2015nonlinear} requires isolated exponentially stable equilibria, not merely attractive invariant sets.
\end{remark}

The parameters $\epsilon_x$ and $\epsilon_\mu$ can be interpreted as the inverse of the estimator \emph{bandwidth} relative to the robot dynamics. For consensus-based estimators tracking time-varying signals, the convergence rate is $O(\lambda_2(L)/\epsilon)$ and the steady-state tracking error scales as $O(\epsilon r / \lambda_2(L))$, where $r$ bounds the rate of change of the tracked signal \cite{kia2019tutorial, bullo2020lectures}. 

For our system, the centroid estimator tracks $x_i(t)$ which varies at rate $\|\dot{x}_i(t)\| \leq 2v_{\max} =: r_x$, and the ascending direction estimator tracks $\mu_i(t) = \sigma(p_i)x_i$ with rate $\|\dot{\mu}_i(t)\| \leq v_{\max} K D =: r_\mu$. Notably, once the formation controller converges and the swarm reaches the desired shape, i.e.,  $\dot{x}_i(t) \to 0$, the centroid estimator effectively tracks a constant signal. In contrast, the ascending direction signal $\mu_i(t)$ continues to vary as the swarm moves through the scalar field, and its rate of change can increase significantly near the source where the gradient varies rapidly.

Thus, smaller products $\epsilon_x r_x$ and $\epsilon_\mu r_\mu$ (equivalently, slower robot motion or faster estimator updates) reduce tracking errors. Better-connected networks (larger $\lambda_2(L)$) permit faster robot motion while maintaining acceptable performance. The ordering $\epsilon_x < \epsilon_\mu$ ensures the centroid estimate converges before being used in the ascending direction computation.

\begin{remark}[Initialization]
To avoid potential singularities when $\|\mu_c^i\|$ is small, the robots' motion may be initiated only after the estimators approach their steady-state values, as determined by their convergence rates.
\end{remark}

\section{Source-seeking with constant-speed 2D unicycle robots}
\label{sec: uni}

We continue our analysis with a robot swarm consisting of 2D unicycle robots traveling at a constant common speed, modeled as in \eqref{eq: ud}. For the sake of conciseness, in this section, we will show that the tracking of $L_\sigma$ can be achieved exponentially fast by the unicycles so that it can be integrated seamlessly in the architecture of Figure \ref{fig: archi}, of course, accounting for the accumulated errors as we have done for system \eqref{eq: sf}.

While we will consider the direction of $L_\sigma$, technically serving as the guiding vector for the robot swarm, it differs in two significant ways from other guiding vector fields \cite{kapitanyuk2017guiding,yao2021singularity,yao2022guiding}. Firstly, the field constructed from $L_\sigma(p_c,x)$ not only depends on the point $p_c$ but also on the deployment $x$, which generally varies during the initial time and can be deliberately morphed for maneuvering purposes. Moreover, during this initial transitory, the centroid $p_c$ may deviate until all the unicycles are well aligned with the desired direction. Secondly, since the signal $\sigma$ is unknown, the robot swarm cannot determine the rate of change of $L_\sigma$ throughout the field. As a result, it cannot be compensated for by a feed-forward term in the control action $\omega_i$ in \eqref{eq: ud} \cite{kapitanyuk2017guiding,yao2021singularity,yao2022guiding}. 

To address these challenges, we will consider first the always-ascending $L^1_\sigma$ for the analysis and subsequently analyze the worst case scenario by considering $L_\sigma$ as a bounded deviation from the ideal direction $L^1_\sigma$. Let us define the set $U := \{p \in \mathbb{R}^{2N} \, | \, L^1_\sigma(p) \neq 0\}$ and introduce $m_d: U \to \mathbb{S}^1$ as
\begin{equation}
\label{eq: md}
m_d(p) = \frac{L^1_\sigma(p)}{\|L^1_\sigma(p)\|} = \begin{bmatrix}\cos(\alpha_d(p)) & \sin(\alpha_d(p))\end{bmatrix}^\top,
\end{equation}
where $\alpha_d \in \mathbb{T}$ and $m_d(p)$ is the \emph{vector field} that must be tracked at $p_c$ to approach the source of the scalar field $\sigma$.

\subsection{Aligning a team of 2D unicycle robots to the vector field of ascending directions}
\label{sec: geofield}


The objective of this section is to design $\omega_i, \forall i\in\mathcal{V}$ in (\ref{eq: ud}) such that $m_i(\alpha_i) = \begin{bmatrix}\cos\alpha_i & \sin\alpha_i \end{bmatrix}^\top \in \mathbb{S}^1$ aligns with $m_d$ sufficiently close. To quantify this alignment, we define the signed angle between $m_d$ and $m_i$ as
\begin{equation*}
\delta_i(p,\alpha_i) :=  \operatorname{atan2}\left((Em_i)^\top m_d,\, m_i^\top m_d\right) \in \mathbb{T},
\end{equation*}
with $E$ being a $+\frac{\pi}{2}$ rotation matrix. The function $\delta_i$ is defined wherever $m_d$ is defined and is continuously differentiable $(C^1)$ whenever $m_i(\alpha_i) \neq - m_d(p)$. Note that the time derivative of $\delta_i$ leads to
\begin{equation} \label{eq_deltadot}
\dot\delta_i = \dot\alpha_i - \dot\alpha_d = \omega_i - \omega_d,
\end{equation}
where $\omega_d := \|\dot m_d\|$, i.e., $\dot m_d = -\omega_d E m_d$.

Our goal is to achieve $\lim_{t\to\infty}\delta_i(t) = 0$, or at least to bound the tracking error effectively. To this end, we adopt the strategy proposed in \cite{kapitanyuk2017guiding}, which involves using a proportional controller but this time without a feed-forward term, i.e.,
\begin{equation} \label{eq_omegai}
    \omega_i(t) =  -\kappa_\gamma\delta_i(t),
\end{equation}
where $\kappa_\gamma \in\mathbb{R}^+$ is a control gain.

\begin{lemma}
\label{lem: x}
Let each robot $i \in \mathcal{V}$ be modeled as in (\ref{eq: ud}) with $\omega_i$ in \eqref{eq_omegai}. Then, the following bound holds
\begin{equation} \label{eq: x_bound}
\|x_i(t) - x_i(0)\| \leq \frac{N-1}{N} \left(\frac{2\pi}{\kappa_\gamma}\right), \forall i \in \mathcal{V}.
\end{equation}
\end{lemma}
\begin{proof}
See Subsection \ref{proof: lem_x} in the Appendix.
\end{proof}

Considering the control law \eqref{eq_omegai}, Lemma \ref{lem: x} provides a way to measure the \emph{deformation} of the deployment concerning its initial condition. In particular, a larger gain $\kappa_\gamma$ results in smaller eventual deformation. Hence, if $x(0)$ is non-degenerated, then $x(t) \to x^*$, where $x^*\in\mathbb{R}^{2N}$ is non-degenerated for a sufficiently large $\kappa_\gamma$. The required magnitude of $\kappa_\gamma$ can be estimated using the minimum eigenvalue of the deployment's covariance matrix $P(x)$. Note that we have normalized the speed of the unicycle to $1$ in (\ref{eq: ud}); therefore, depending on the robot speed or length units, the deformation is directly proportional to such quantity.

Substituting \eqref{eq_omegai} into \eqref{eq_deltadot} yields
\begin{equation}
\label{eq: wp}
\dot{\delta}_i(t) = \omega_i(t) - \omega_d(t) = -\kappa_\gamma\delta_i(t) - \omega_d(t),
\end{equation}
however, the robot swarm does not have knowledge of $\omega_d$, as the signal $\sigma$ is unknown in the source-seeking problem. We will design $\kappa_\gamma$ by treating the upper bound of $|\omega_d(p)|$ as a non-vanishing bounded perturbation. However, $|\omega_d(p)|$ is not globally bounded, as the vector field $m_d(p)$ is not well-defined for $p \in \{p \, : \, p_c = p_\sigma \}$ because $\nabla\sigma(p_\sigma) = 0$. Furthermore, in a neighborhood of such a set, $\omega_d$ changes rapidly; e.g., when $p_c$ crosses $p_\sigma$ the vector field $m_d(p)$ spins $\pi$ radians \emph{instantaneously}. Nevertheless, focusing our analysis on trajectories that remain outside this critical neighborhood still ensures that the robot swarm approaches the source, as required by Problem \ref{prob: ss}. In the following technical result, we prove for such trajectories that an upper bound for $|\omega_d(p)|$ always exists.

\begin{lemma}
\label{lem: Omegad}
Let each robot $i \in \mathcal{V}$ be modeled as in (\ref{eq: ud}) with $\omega_i$ in \eqref{eq_omegai} and $x(0)$ non-degenerated.
Then, there exists a constant $\kappa > 0$ such that, if $\kappa_\gamma \ge \kappa$,  for any $\varepsilon > 0$ satisfying $\|\nabla\sigma(p_c(t))\| \ge \varepsilon$, it holds that
$|\omega_d(p(t))| \le \Omega_d(\varepsilon)$, $\forall t \ge 0$,
where $\Omega_d(\varepsilon) > 0$.
\end{lemma}
\begin{proof}
See Subsection \ref{proof: lem_omegad} in the Appendix.
\end{proof}

Similar to the result in \cite[Lemma 9.2]{khalil2015nonlinear} on bounded disturbances, having $|\omega_d(p)|$ uniformly bounded by Lemma \ref{lem: Omegad}, the following technical result shows how to design $\kappa_\gamma$ to ensure that all $\delta_i$ remain within a bounded interval after some finite time; i.e., the team of 2D unicycles aligns with the vector field $m_d(p)$ sufficiently close.

\begin{prop}
\label{prop: gamma}
	Let each robot $i \in \mathcal{V}$ be modelled as in (\ref{eq: ud}) with $\omega_i$ in \eqref{eq_omegai} and $x(0)$ non-degenerate. Then, for any given angle $\gamma \in (0, \frac{\pi}{2})$, $\epsilon > 0$, and time $T \in \mathbb{R}^+$ there exist a constant $\kappa > 0$ such that if $\kappa_\gamma > \kappa$ and $\|\nabla\sigma(p_c(t))\| > \epsilon, \forall t\in[0, T)$ then $|\delta_i(T)| < \gamma, \forall i\in\mathcal{V}$. Furthermore, this bound remains valid $\forall t\in[T, T^*)$, where $T^* \geq T$ denotes the first time such that $\|\nabla\sigma(p_c(T^*))\| \leq \epsilon$.
\end{prop}
\begin{proof}
See Subsection \ref{proof: lem_gamma} in the Appendix.
\end{proof}

\subsection{Solution to the source-seeking problem for a team of 2D unicycle robots with a constant common speed}

All the technical results in the previous subsections are not so relevant in practice, but they are just for formal analysis. Indeed, having $x(t)$ degenerated in real applications is practically impossible due to disturbances or error measurements. Considering a signal as defined in to Definition \ref{signal}, the variation of $\nabla \sigma (p_c)$ due to the robots' movement is sufficiently small and, in any case, can be bounded as shown in Lemma \ref{lem: Omegad}, at least far enough from the source. However, once the centroid of the swarm is close to the source, it can change rapidly, and they might not track the guiding vector field. Nonetheless, the robot swarm will repeatedly fall into a neighborhood close to the source while the deployment $x(t)$ converges exponentially fast to a fixed one close to $x(0)$. Note that, unlike the single-integrator controller, the shape of the deployment is not explicitly under control. As a result, although the final formation remains non-degenerate, its exact shape is arbitrary.

Before proceeding to the formal proof, we provide a brief and intuitive overview of the main ideas involved. The goal is to show that, for any prescribed $\epsilon > 0$, one can choose a sufficiently large gain $\kappa_\gamma$ so that the centroid eventually enters and remains within the region $\mathcal{B} := \{ y \in \mathbb{R}^2 : \| y - p_\sigma \| < \epsilon \}.$

A direct application of a standard convergence theorem is not feasible. Indeed, the unicycles move at constant speed along a direction determined by a heading angle, and when they are very close to the source, this angle may rotate arbitrarily fast. This behaviour prevents a straightforward compactness or monotonicity argument.

To overcome this difficulty, we introduce a “buffer zone” $\mathcal{A} \subseteq \mathcal{B}$. Although we may lose precise control of the centroid once it lies inside $\mathcal{A}$, we retain sufficient control in the annular region $\mathcal{B} \setminus \mathcal{A}$. This ensures that, even if the centroid leaves $\mathcal{A}$, it cannot escape from $\mathcal{B}$. Consequently, the centroid remains confined within $\mathcal{B}$ once it has reached $\mathcal{A}$ for the first time. Thus, to establish the result, we just need to show that the centroid eventually reaches region $\mathcal{A}$. Keeping this picture in mind, we proceed to the proof of the theorem.

\begin{theorem}
\label{th: uni}
	Let $\sigma$ be a signal according to Definition \ref{signal}. Consider a robot swarm of $N$ constant-speed 2D unicycle robots modelled by \eqref{eq: ud}, with a non-degenerate initial deployment $x(0)$, and $\|p_c(0)  - p_\sigma\| > 0$. Then, given an $\epsilon>0$ there exists a gain $\kappa_\gamma > 0$ such that the control law $\omega_i = -\kappa_\gamma \delta_i, \forall i \in \mathcal{V}$, provides a solution to Problem \ref{prob: ss}.
\end{theorem}
\begin{proof}
	Fix $\epsilon > 0$. We will show that there exists a positive gain $\kappa_\gamma$ for the angular velocity controller of the unicycles such that the centroid trajectory $p_c(t)$ stays within the open set $\mathcal{B}:=\{y\in\mathbb{R}^2 \, : \, \|y - p_\sigma \| < \epsilon \}$, which is precisely the solution to Problem \ref{prob: ss}.
    
    By selecting $\kappa_1$ as in Lemma \ref{lem: x}, we can ensure that there exists a compact set $\mathcal{X}\subset\mathbb{R}^{2N}$ centered at $x(0)$ such that if $\kappa_\gamma > \kappa_1$, then $x(t) \in \mathcal{X}$ for all $t \geq 0$, i.e. the deployment remains non-degenerate and inside the compact set $\mathcal{X}$. Let
    $$
    \theta(t) := \theta(L^1_\sigma(p_c(t), x(t)), \nabla\sigma(p_c(t)))
    $$
    be the angle between the direction $L^1_\sigma(p_c(t), x(t))$ and the gradient $\nabla\sigma(p_c(t))$. From Proposition \ref{prop: L_sigma} we know
	$$
	\frac{L^1_\sigma(p_c(t), x(t))^\top\nabla\sigma(p_c(t))}{\|L^1_\sigma(p_c(t), x(t))\| \|\nabla\sigma(p_c(t))\|} = \cos(\theta(t)) > 0,
	$$
    while whenever $\kappa_\gamma > \kappa_1$, the function $\cos(\theta(t))$, seen as a continuous function of the direction $L^1_\sigma(p_c(t), x(t))$ and the gradient $\nabla\sigma(p_c(t))$, takes values inside the compact set $\mathbb{S}^1 \times \mathcal{X}$ along the trajectory. In particular, we can find $H^* > 0$ such that $0 < H^* < \cos (\theta(t))$ along the trajectory. This implies that there exists a constant $\gamma > 0$ such that
    
    \begin{equation}
        |\theta(t)| < \frac{\pi}{2} - 2\gamma \label{eq: uniform-control}
    \end{equation}
    for all $t > 0$, as long as $p_c(t) \neq p_\sigma$. Observe that this $\gamma$ is valid for any choice of $\kappa_\gamma > \kappa_1$. We fix such a $\gamma$ going forward.

    Next, let us choose $\epsilon^*>0$ small enough such that the set $\mathcal{A} := \{y\in \mathcal{B} \, : \, \left\| \nabla\sigma(y) \right\|\leq \epsilon^*  \} \subset \mathcal{B}$ satisfies $d = \operatorname{dist}(\mathcal{A},\mathbb{R}^2\setminus\mathcal{B}) > 0$. Observe that such a $\epsilon^*$ exists from the continuity of $ \nabla\sigma$. Apply Proposition \ref{prop: gamma} with the chosen $\gamma$, $\epsilon^*$ and $T = d/2$, so that there exists a gain $\kappa_2>0$ such that if $\kappa_\gamma > \kappa_2$, then $|\delta_i(t)| < \gamma$ after the centroid has been $T$ units of time outside of $\mathcal{A}$, provided it never re-entered it.
    We claim that choosing 
    $$
    \kappa_\gamma \geq \operatorname{max}\{\kappa_1, \kappa_2\}
    $$ 
    works. 
    
    We argue in two steps. First, we show that after entering the region $\mathcal{A}$, the centroid cannot leave the region $\mathcal{B}$. Secondly, we argue that there exists a finite time $T^*$ such that the centroid enters region $\mathcal{A}$, which will finish the proof. 

    Suppose that at some time the centroid lies inside the region $\mathcal{A}$. If at some later time the centroid leaves $\mathcal{A}$, we have two alternatives. First, it re-enters $\mathcal{A}$ before a time $T$ has elapsed. Since $T = d/2$ and $\norm{\dot{p}_c(t)} \leq 1$, the centroid did not have time to leave $\mathcal{B}$, in which case, we are back inside $\mathcal{A}$ and repeat this argument if necessary. Alternatively, suppose that after a time $T$ has passed, we are yet to re-enter $\mathcal{A}$. As before, this time is not enough for the centroid to travel outside of $\mathcal{B}$, so we are still inside $\mathcal{B}$. Furthermore, since we assumed we never re-entered $\mathcal{A}$, our previous choice guarantees that $|\delta_i(t)| < \gamma$ until we re-enter $\mathcal{A}$, which together with \eqref{eq: uniform-control} gives that we can find a constant $c > 0$ such that
    \begin{equation*}
        \begin{bmatrix}\cos\alpha_i(t) & \sin\alpha_i(t) \end{bmatrix}\nabla\sigma(p_c(t)) > c \epsilon^* > 0
    \end{equation*}
    for all $i\in\mathcal{V}$ after $T$ units of time since we left $\mathcal{A}$ and until $p_c(t)$ re-enters $\mathcal{A}$. Hence, until we re-enter $\mathcal{A}$ we have
    \begin{align}
    \frac{\mathrm{d}}{\mathrm{dt}}\sigma(p_c(t)) 
    &= \left(\frac{\mathrm{d}p_c(t)}{\mathrm{dt}}\right)^\top \nabla\sigma(p_c(t)) \nonumber \\
    &= \frac{1}{N}\sum_{i=1}^N 
    \begin{bmatrix}\cos\alpha_i(t) \\ \sin\alpha_i(t) \end{bmatrix}^\top
    \nabla\sigma(p_c(t)) > c\epsilon^*. 
    \label{eq: uniform-lower-bound}
    \end{align}

    In particular, since at this point the centroid is still inside the sublevel set $\mathcal{B}$, and $\sigma(p_c(t))$ is now increasing through the trajectory, it cannot leave this sublevel set. This proves that once the centroid enters $\mathcal{A}$, it remains trapped in the region $\mathcal{B}$.

    To conclude, we show that the centroid enters the region $\mathcal{A}$ in finite time. The reader may observe that the following argument is a direct Lyapunov-type reasoning and may be viewed as a particular instance of LaSalle's invariance principle applied to $\sigma(p_c(t))$. We present it explicitly rather than invoking LaSalle directly because the control over the derivative of $\sigma$ is only guaranteed outside $\mathcal{A}$, and we wish to make the argument constructive within this region.

    Suppose by contradiction that $p_c(t) \notin \mathcal{A}$ for all $t \ge 0$. From the previous argument, once $p_c(t)$ has remained outside $\mathcal{A}$ for a time greater than $T$, the derivative $\frac{\mathrm{d}}{\mathrm{d}t}\sigma(p_c(t))$ admits a uniform positive lower bound \eqref{eq: uniform-lower-bound}. In particular, integrating in time for $t > T$ yields
    \begin{equation*}
    \sigma(p_c(t)) \ge \sigma(p_c(T)) + c \epsilon^* (t - T),
    \end{equation*}
    which diverges to $+\infty$ as $t \to \infty$. This contradicts the global boundedness of $\sigma$, which shows that there must exist a finite time $T^*$ such that $p_c(T^*) \in \mathcal{A}$. This finishes the proof.
\end{proof}

\begin{remark}
	Note that by definition, the swarm of unicycle robots moves continuously, i.e, they move along orbits eventually of equal radii determined by the gain $\kappa_\gamma$.
\end{remark}

Indeed, tracking the actual $L_\sigma$ and not $L^1_\sigma$ by the robot swarm makes a deviation characterized by Proposition \ref{prop: L_sigma}. However, as in\cite[Theorem 1]{acuaviva2023resilient}, the solution to Problem \ref{prob: ss} by Theorem \ref{th: uni} is still valid for a given $||p_c(t) - p_\sigma|| < \epsilon$ with an extra bound for $D$ considering the bound of the Hessian $M$ and $\epsilon$ in Proposition \ref{prop: D_max}. Of course, such a bound on $D$ will be more relaxed for deployments considering $L^1_\sigma$ parallel to the gradient, i.e., for bigger deviations of $L_\sigma$ due to a big $D$, we could still have $L_\sigma(p_c, x)^\top \nabla\sigma(p_c)> 0$ for longer time or convergence sets much closer to the source $p_\sigma$.

\begin{remark}
	If the estimation of the ascending direction is obtained via the two consensus algorithms described in Figure \ref{fig: archi}, then the bounds given in Theorem \ref{th: uni} will be even more conservative since the centroid and ascending directions calculated distributively by the robots are not exactly the eventual asymptotic values considered in the main results.
\end{remark}


\section{Numerical simulations}
\label{sec: sce}
All the numerical validations and simulations can be found in the following repository \url{https://github.com/Swarm-Systems-Lab/source_seeking}.

\begin{figure}
\centering
\includegraphics[trim={0cm 0cm 0cm 0cm}, clip, width=1\columnwidth]{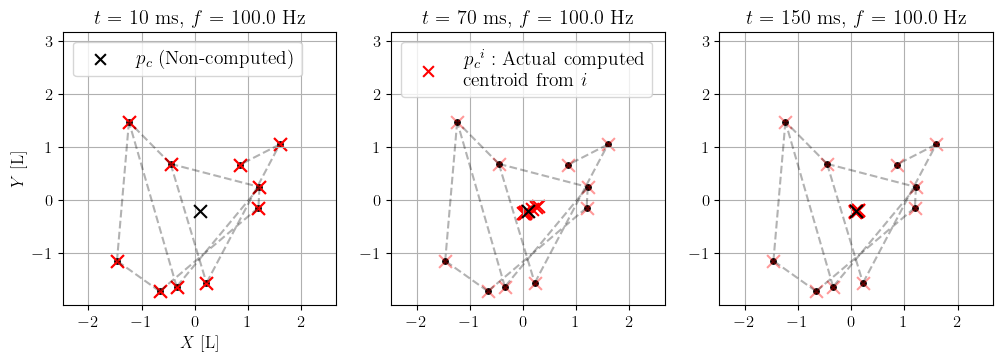}
	\caption{Three snapshots from the algorithm in Corollary \ref{cor: centroid} that estimates the centroid of a robot swarm distributively. We can note the asymptotic convergence to the actual value of the centroid (X in black color), the X's in red color are the values of $p_c^i(t) = p_i - \hat{x}_i(t)$. Each robot starts with $\hat{x}_i(0) = 0$ which is equivalent to start from their own positions (X's in light red color). Dashed lines indicate neighboring robot pairs. 
}
\label{fig: centroid}
\end{figure}

\begin{figure}
\centering
\includegraphics[trim={0cm 0cm 0cm 0cm}, clip, width=1\columnwidth]{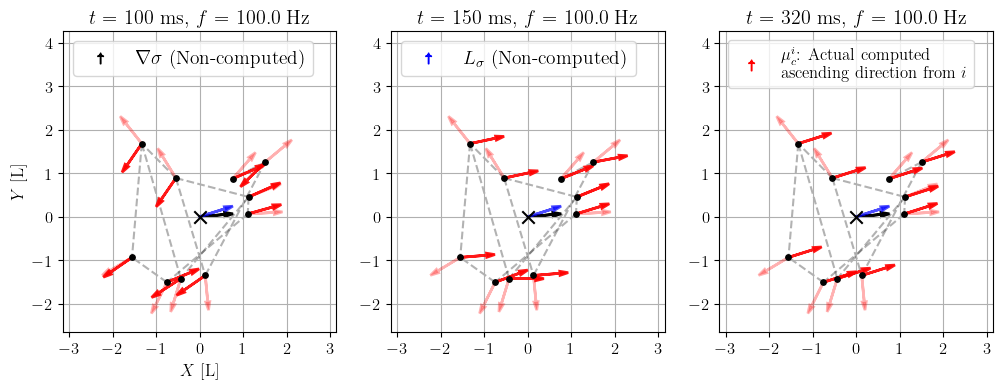}
	\caption{Three snapshots from the algorithm in Corollary \ref{cor: ascending} that estimates the ascending direction at the centroid $L_\sigma/\|L_\sigma\|$ (blue arrow). The initials $\mu_c^i(0) = \mu_i(0) - \hat \mu_i(0)$, with $\hat\mu_i(0) = 0$, are in light red color, and eventually we can see how $\mu_c^i/\|\mu_c^i\|$ aligns with the ascending direction $L_\sigma$. Since only directional information is relevant, all vectors shown in this figure are normalized to a common magnitude. Dashed lines indicate neighboring robot pairs.
}
\label{fig: ascending}
\end{figure}

\subsection{Distributed estimation of the swarm centroid and ascending direction}
We provide numerical validation of Corollary \ref{cor: centroid} in Figure \ref{fig: centroid}, and of Corollary \ref{cor: ascending} in Figure \ref{fig: ascending}. The simulations are based on the following set of edges 
$\scalemath{0.78}{\mathcal{E} =\{(0, 3), (0, 9), (3, 6), (0, 8), (8, 5), (6, 7), (1, 4), (4, 2), (8, 1), (9, 2), (5, 9)\}}$,
which corresponds to a graph with algebraic connectivity $\lambda_2 = 0.26$. In Figure \ref{fig: centroid}, each robot asymptotically estimates the centroid of the swarm $p_c$ relative to its own reference frame centered at $p_i$, i.e., the centroid-relative coordinates $-x_i$. In Figure \ref{fig: ascending}, a similar estimation process is shown, but for the centroid of the directions $\mu_i$, denoted by $\mu_c$, which is parallel to the collective ascending direction $L_\sigma$.

\begin{figure*}[t!]
	\centering
	\includegraphics[trim={0cm 0.2cm 0cm 0.5cm}, clip, width=2\columnwidth]{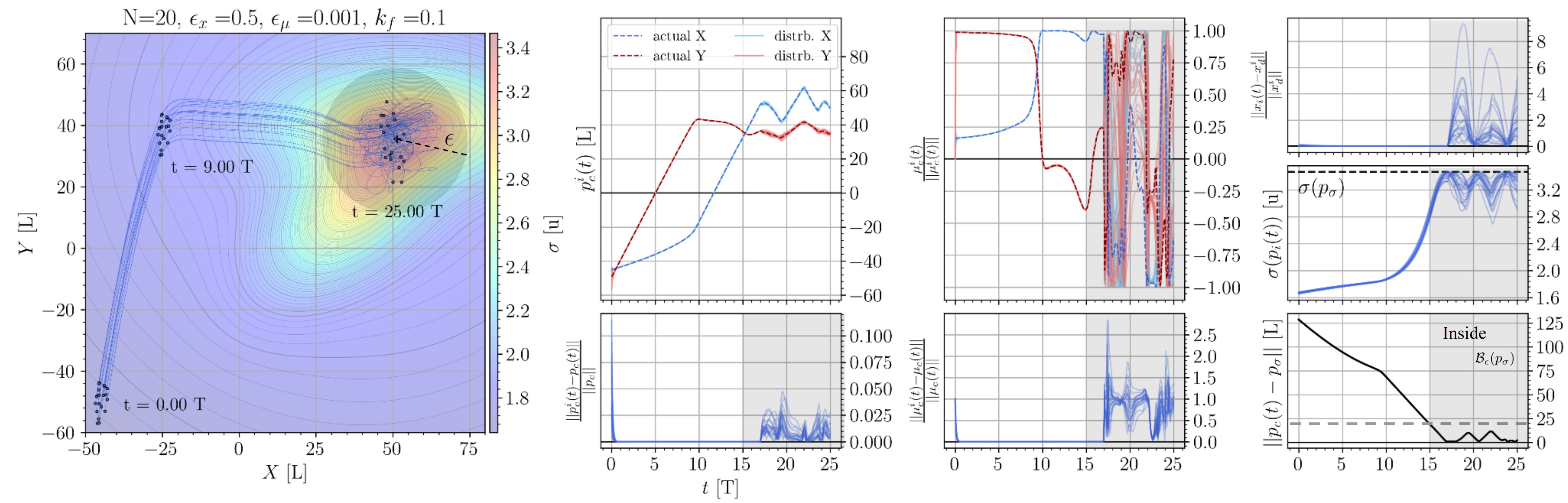}
		\caption{A swarm of $30$ single-integrator robots (blue dots), modeled by the slow-fast closed-loop system \eqref{eq: sf}. In the two middle columns: the right plot shows the estimated centroid $p_c^i(t) = p_i(t) - \hat{x}_i(t)$, with the corresponding mean deviation error displayed below; the left plot shows the evolution of the normalized estimated direction $\mu_c^i(t)/\|\mu_c^i(t)\|$, also accompanied by its mean deviation error. In the rightmost column, from top to bottom: the mean deviation error between the actual formation $x(t)$ and the desired one $x_d$; the scalar signal readings of the robots (blue lines), with the source value indicated by a black dashed line; and the norm of the relative position between the robots and the source. Grey regions in the plots indicate time intervals during which $p_c(t)$ is inside the $\epsilon$-neighborhood of the source, $\mathcal{B}_\epsilon(p_\sigma) = \{p \in \mathbb{R}^m : \|p - p_\sigma\| < \epsilon\}$.
		}
	\label{fig: si}
\end{figure*}

\begin{figure}
	\centering
	\includegraphics[trim={0cm 0cm 0cm 0cm}, clip, width=1\columnwidth]{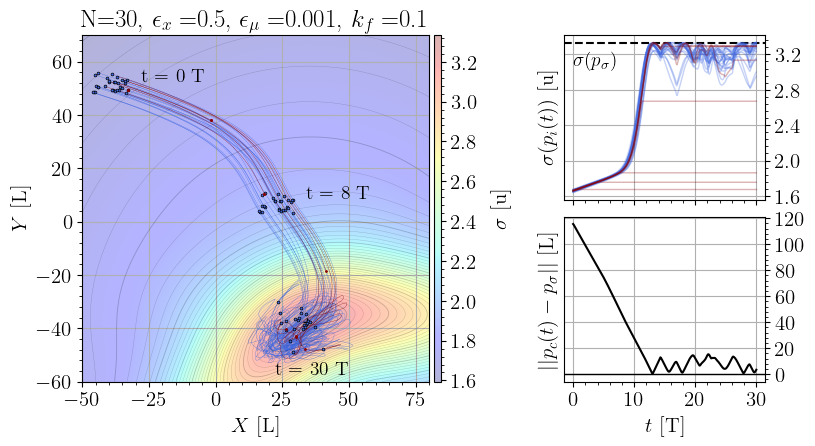}
	\caption{A swarm of $30$ single-integrator robots, modeled by the slow-fast closed-loop system \eqref{eq: sf}. During the mission, $8$ robots are deliberately disconnected at different time steps (red dots) to demonstrate that the remaining \emph{alive} robots (blue dots) successfully reach the source, highlighting the swarm's resiliency. On the right, from top to bottom, the signal readings of the robots are shown, with the source value indicated by a black dashed line. Below, the norm of the relative position between the robots and the source; disconnected robots are excluded from the computation of $p_c(t)$.}
	\label{fig: si_missing}
\end{figure}

\begin{figure}
\centering
\includegraphics[trim={0cm 0cm 0cm 0cm}, clip, width=0.98\columnwidth]{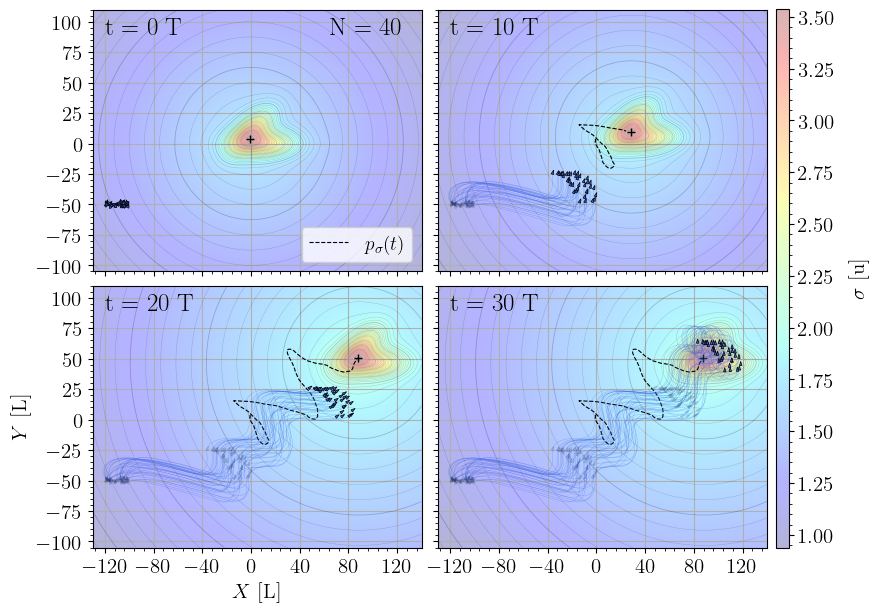}
\caption{A swarm of $40$ 2D unicycle robots (blue triangles) travelling at constant speed tracks a \emph{dynamic scalar field source}, whose trajectory is shown by the dashed black line $p_\sigma(t)$. The swarm eventually reaches the source and begins to orbit around it.}
\label{fig: uni2d}
\end{figure}

\subsection{Distributed source-seeking with single integrators}
In Figure \ref{fig: si}, we deploy $20$ single-integrator robots modeled as in \eqref{eq: sf} and measuring the non-convex signal
\begin{align}
	\scalemath{0.85}{\sigma(p) =} 
	& \scalemath{0.85}{ 
	\; k \frac{x_\sigma(p)}{\|x_\sigma(p)\|} - Q_a (x_\sigma(p) - a) \exp \left\{ (x_\sigma(p) - a)^\top Q_a (x_\sigma(p) - a)\right\}} \nonumber \\
	& \quad\scalemath{0.85}{
	- Q_b (x_\sigma(p) - b) \exp \left\{ (x_\sigma(p) - b)^\top Q_b (x_\sigma(p) - b)\right\}},
	\label{eq: sigma_sim}
\end{align}
where
$x_\sigma(p) := (p - p_\sigma) / \delta$, $\delta = 15$, 
$k=0.04$,
$a = \left[\begin{smallmatrix}1\\0\end{smallmatrix}\right]$,
$b = \left[\begin{smallmatrix}0\\-1.5\end{smallmatrix}\right]$,
$p_\sigma = \left[\begin{smallmatrix}40\\40\end{smallmatrix}\right]$, 
$Q_a = 0.9\left[\begin{smallmatrix} \frac{1}{\sqrt{30}} & 0 \\ 1 & 0 \end{smallmatrix}\right]$,
$Q_b = - A^\top S A$, with 
$A = \frac{1}{\sqrt{2}}\left[\begin{smallmatrix} 1 & -1 \\ 1 & 1 \end{smallmatrix}\right]$ and
$S = 0.9\left[\begin{smallmatrix} 1 & 0 \\ 0 & \frac{1}{\sqrt{15}} \end{smallmatrix}\right]$.
Both, the centroid and the ascending direction are calculated distributively, using $\epsilon_x = 0.5$ and $\epsilon_\mu = 0.001$. The underlying formation controller consider $x_d$ as the initial formation and try to preserve it with $k_f = 0.1$. This initial formation is a uniform rectangular distribution. Eventually, the centroid swarm gets so close to the source that $L_\sigma$ becomes unreliable, i.e., the centroid entered the $\epsilon$-ball in Problem \ref{prob: ss}.

In Figure \ref{fig: si_missing}, the same scalar field in \eqref{eq: sigma_sim} is considered, but this time $\delta = 20$, $k = 0.06$,
$a = \left[\begin{smallmatrix}1.5\\0\end{smallmatrix}\right]$,
$b = \left[\begin{smallmatrix}-1\\-1\end{smallmatrix}\right]$, and
$p_\sigma = \left[\begin{smallmatrix}35\\-35\end{smallmatrix}\right]$. This time we deploy $30$ single-integrator robots, also modeled as in \eqref{eq: sf}, but to validate the resilience of our source-seeking algorithm, we deliberately disconnect a robot every $3.75$ T, so that at the end of the mission eight robots have been disconnected from the network while ensuring that the alive robots keep connected. Even though with the presence of dead robots, the alive agents eventually find the source of the scalar field.

\subsection{Source-seeking with unicycles and a dynamic source}
In Figure \ref{fig: uni2d}, we illustrate the effectiveness of Theorem \ref{th: uni} using 2D unicycle robots traveling at constant speed to track a dynamic scalar field source. Although the agents begin with arbitrary headings, they converge to a common direction closely aligned with the one defined by $L_\sigma$. Notably, once the swarm centroid approaches the source, the robots begin to describe close orbits, thereby bounding the distance between the centroid and the source.

\section{Conclusion}
\label{sec: con}
We have presented a fully distributed methodology to solve the source seeking problem with robot swarms by dynamically estimating both the centroid and an ascending direction using consensus-based algorithms, assuming single-integrator dynamics. Unlike previous methods, our approach does not require strict formation constraints, allowing for deformable swarm shapes and enabling shape morphing during motion. For unicycle dynamics, we have demonstrated a centralized solution that remains effective despite more restrictive kinematic constraints. Theoretical results were supported by extensive simulations, validating the convergence properties and robustness of our approach under a variety of conditions.

Future work will deepen the analysis of the distributed system, quantifying the impact of imperfect measurements, and the trade-offs between estimation speed, agent velocity, and scalar field smoothness to ensure more accurate convergence. These insights will guide the specification of practical requirements, helping us to implement the algorithm in the real world using custom hardware designed to meet the computational and communication demands identified by our analysis.
Furthermore, we aim to extend our results to 3D unicycle robots, where geometric SO(3)/SE(3)-based control laws will be developed to align spatially constrained agents, extending our framework to more complex environments.


\begin{appendix}

\subsection{Proof of Proposition \ref{prop: single_failure_weyl}} \label{proof: single_failure_weyl}

\begin{proof}
Let $D = \max_{i \in \mathcal{V}} \|x_i\|$ and $D^{(-j)} = \max_{i \in \mathcal{V} \setminus \{j\}} \|x_i^{(-j)}\|$. Since the original deployment satisfies $\sum_{i \in \mathcal{V}} x_i = 0$, we have $\sum_{i \neq j} x_i = -x_j$. The new barycentric coordinates are
\begin{equation*}
x_i^{(-j)} = x_i - \frac{1}{N-1}\sum_{k \neq j} x_k = x_i + \frac{x_j}{N-1}, \quad \forall i \neq j.
\end{equation*}
By the triangle inequality,
\begin{equation}
\scalemath{0.8}{
\|x_i^{(-j)}\| = \left\|x_i + \frac{x_j}{N-1}\right\| \le \|x_i\| + \frac{\|x_j\|}{N-1} \le \frac{N}{N-1}D,
}
\label{eq:D_bound}
\end{equation}
so $D^{(-j)} \le \frac{N}{N-1}D$.

Define $A = \sum_{i=1}^N x_i x_i^\top$ and $A^{(-j)} = \sum_{i \neq j} x_i^{(-j)} (x_i^{(-j)})^\top$. Then
\begin{align*}
\scalemath{0.8}{A^{(-j)}} &\scalemath{0.8}{= \sum_{i \neq j} \left(x_i + \frac{x_j}{N-1}\right)\left(x_i + \frac{x_j}{N-1}\right)^\top} \nonumber \\
&\scalemath{0.8}{= \sum_{i \neq j} x_i x_i^\top + \frac{1}{N-1}\sum_{i \neq j} \left(x_i x_j^\top + x_j x_i^\top\right) + \frac{x_j x_j^\top}{(N-1)^2}(N-1)}.
\end{align*}
Using $\sum_{i \neq j} x_i = -x_j$ and $A_{-j} := \sum_{i \neq j} x_i x_i^\top = A - x_j x_j^\top$:
\begin{align*}
\scalemath{0.8}{A^{(-j)}} &\scalemath{0.8}{= A_{-j} - \frac{2x_j x_j^\top}{N-1} + \frac{x_j x_j^\top}{N-1}} \\
&\scalemath{0.8}{= A - x_j x_j^\top - \frac{x_j x_j^\top}{N-1} = A - \frac{N}{N-1}x_j x_j^\top.}
\end{align*}

The normalized second-moment matrices are
\begin{equation*}
S(x) = \frac{A}{ND^2}, \quad S(x^{(-j)}) = \frac{A^{(-j)}}{(N-1)(D^{(-j)})^2}.
\end{equation*}
Their difference is
\begin{equation*}
S(x) - S(x^{(-j)}) = \frac{A}{ND^2} - \frac{A - \frac{N}{N-1}x_j x_j^\top}{(N-1)(D^{(-j)})^2}.
\end{equation*}
Using $(D^{(-j)})^2 \le \frac{N^2}{(N-1)^2}D^2$ from \eqref{eq:D_bound}, and bounding each term with $\|x_j x_j^\top\| \le D^2$ and $\|A\| \le ND^2$, we obtain
\begin{equation*}
\|S(x) - S(x^{(-j)})\| \le \frac{4}{N-1}.
\end{equation*}
The result follows from Weyl's inequality for symmetric matrices.
\end{proof}

\subsection{Proof of Lemma \ref{lem: x}} \label{proof: lem_x}

\begin{proof}
Let us define $\delta_{ij} := \alpha_i - \alpha_j$, which is the oriented angle from the heading of robot $i$ towards the heading of robot $j$, so that $\delta_{ij} \in (-2\pi,2\pi)$ and if $\delta_{ij} = 0$, then robots $i$ and $j$ share the same heading. Note that $\delta_{ij} = \delta_i - \delta_j$; hence
\begin{equation}
	\dot\delta_{ij} = \omega_i - \omega_j = -\kappa_\gamma(\delta_i - \delta_j) = -\kappa_\gamma\delta_{ij}, \nonumber
\end{equation}
whose trajectory can be described by $\delta_{ij}(t) = \delta_{ij}(0)e^{-\kappa_\gamma t}$,
thus we have that $\lim_{t\to\infty}\delta_{ij} = 0$. Now we are ready to (over)estimate the deformation of the relative positions concerning their initial states. We start by considering
\begin{align}
	& \scalemath{0.8}{\dot p_{ij} = \dot p_i - \dot p_j = \begin{bmatrix}\cos\alpha_i - \cos\alpha_j \\ \sin\alpha_i - \sin\alpha_j \end{bmatrix} = 2\sin\left(\frac{\alpha_i - \alpha_j}{2}\right)\begin{bmatrix}-\sin\left(\frac{\alpha_i + \alpha_j}{2}\right) \\ \cos\left(\frac{\alpha_i + \alpha_j}{2}\right)\end{bmatrix}} \nonumber \\
	& \scalemath{0.8}{= 2\sin\left(\frac{\delta_{ij}}{2}\right)\begin{bmatrix}-\sin\left(\frac{\alpha_i + \alpha_j}{2}\right) \\ \cos\left(\frac{\alpha_i + \alpha_j}{2}\right)\end{bmatrix} = 2\sin\left(\frac{\delta_{ij}(0)e^{-\kappa_\gamma t}}{2}\right)\begin{bmatrix}-\sin\left(\frac{\alpha_i + \alpha_j}{2}\right) \\ \cos\left(\frac{\alpha_i + \alpha_j}{2}\right)\end{bmatrix}},\nonumber
\end{align}
which leads to the inequality
\begin{equation}
	\scalemath{0.8}{
	\|\dot p_{ij}(t)\| \leq 2\left|\sin\left(\frac{\delta_{ij}(0)e^{-\kappa_\gamma t}}{2}\right) \right| \leq |\delta_{ij}(0)|e^{-\kappa_\gamma t}}, \nonumber
\end{equation}
Therefore, the relative positions stabilize exponentially fast. In fact, since $p_{ij}(t) = p_{ij}(0) + \int_0^\top \dot p_{ij}(s)\,\mathrm{d}s$
we conclude that
\begin{align}
	& \scalemath{0.8}{\|p_{ij}(t) - p_{ij}(0)\| = \left\| \int_0^\top \dot p_{ij}(s) \, \mathrm{d}s\right\| \leq \int_0^\top \|\dot p_{ij}(s) \| \, \mathrm{d}s \leq} \nonumber \\
	& \scalemath{0.8}{\leq |\delta_{ij}(0)| \int_0^\top e^{-\kappa_\gamma t} \leq |\delta_{ij}(0)| \int_0^\infty e^{-\kappa_\gamma s} \, \mathrm{d}s = \frac{|\delta_{ij}(0)|}{\kappa_\gamma} \leq \frac{2\pi}{\kappa_\gamma}},\nonumber
\end{align}
where we have used $\delta_{ij}(0) \in (-2\pi, 2\pi)$.

Next, we observe that 
$$\scalemath{0.8}{x_i = p_i - \frac{1}{N}\sum_{j\in\mathcal{V}}p_j = \frac{1}{N}\sum_{j\in\mathcal{V}}(p_i - p_j) = \frac{1}{N}\sum_{j\in \mathcal{V} \setminus \{i\}}(p_i - p_j)},$$ 
thus
\vspace{0.3cm}

$\scalemath{0.8}{\|x_i(t) - x_i(0)\| \leq \frac{1}{N} \sum_{j\in\mathcal{V} \setminus \{i\}} \|p_{ij}(t) - p_{ij}(0)\| \leq \frac{N-1}{N} \left(\frac{2\pi}{\kappa_\gamma}\right)}.$
\end{proof}

\subsection{Proof of Lemma \ref{lem: Omegad}} \label{proof: lem_omegad}

Before the main proof, first we need to provide the following technical result.
\begin{lemma}
\label{lem: Lgrad}
Let each robot $i \in \mathcal{V}$ be modeled as in (\ref{eq: ud}) with $\omega_i$ in \eqref{eq_omegai} and $x(0)$ non-degenerated. Then, there exist constant $\kappa > 0$ such that if $\kappa_\gamma \geq \kappa$, it holds that $\|L^1_\sigma(p_c(t), x(t))\| \geq \frac{\lambda_{\text{min}}\{P(x(t))\}}{D^2} \|\nabla\sigma(p_c(t))\|$.
\end{lemma}
\begin{proof}
Because $x(0)$ is non-degenerated, from Lemma \ref{lem: x} we know that there exist $\kappa > 0$ so that if $\kappa_\gamma > \kappa$ then $x(t), \forall t \geq 0$ is non-degenerated. Then, considering the lower Rayleigh quotient bound of $L^1_\sigma(p_c, x)^\top\nabla\sigma(p_c)$ and the standard Cauchy-Schwarz inequality, we have that $\frac{\lambda_{\text{min}}\{P(x)\}}{D^2} \|\nabla\sigma(p_c)\|^2 \leq L^1_\sigma(p_c, x)^\top\nabla\sigma(p_c) \leq \|L^1_\sigma(p_c, x)\|\,\|\nabla\sigma(p_c)\|$, which proves the claim.
\end{proof}

\begin{proof}[Proof of Lemma \ref{lem: Omegad}]
Since $\|m_d\|^2 = 1$, it follows from \cite[Lemma 3.1]{yuri2020motion} that $\omega_d(t) = \dot m_d(t)^\top E m_d(t)$. Consecuently, $|\omega_d(t)| \leq \|\dot m_d(t)\| \, \|m_d(t)\| = \|\dot m_d(t)\|$. Therefore, bounding $|\omega_d(t)|$ reduces to bounding $\|\dot m_d(t)\|$.

From the definition of $m_d$ in \eqref{eq: md}, we have 
\begin{equation}
	m_d = \frac{L_\sigma^1}{\|L_\sigma^1\|} = \frac{\sum_{i=1}^N(\nabla\sigma(p_c)^\top (p_i - p_c)(p_i - p_c))}{\|\sum_{i=1}^N(\nabla\sigma(p_c)^\top (p_i - p_c)(p_i - p_c))\|}. \nonumber
\end{equation}
Calculating the time derivative of one term in the sum gives
\begin{align} \label{eq: dt_sum}
	&\scalemath{0.9}{\frac{\mathrm{d}}{\mathrm{dt}}\left(\nabla\sigma(p_c)^\top (p_i - p_c)(p_i - p_c) \right) =} \nonumber \\
	&\quad\scalemath{0.9}{= \left[(H_\sigma(p_c)\dot p_c)^\top (p_i - p_c) + \nabla\sigma(p_c)^\top (\dot p_i - \dot p_c) \right] (p_i - p_c) } \nonumber \\
	&\qquad\scalemath{0.9}{+ \; \nabla\sigma(p_c)^\top (p_i - p_c) (\dot p_i - \dot p_c).}
\end{align}

Regarding the different terms in \eqref{eq: dt_sum}, from Lemma \ref{lem: x}, we know that $\|p_i - p_c\| = \|x_i\| \leq \frac{2\pi}{\kappa_\gamma} + \|x_i(0)\| \leq M_1$. Additionally, we have that $\|\dot p_i - \dot p_c\| = \|\dot p_i - \frac{1}{N}\sum_{j=1}^N \dot p_j\| \leq 2$, and, indeed, we also have that $\|\dot p_c\| \leq 1$. Finally, we remind that the Hessian and the gradient of $\sigma$ are also bounded by Definition \ref{signal}. Hence, it follows that
\begin{equation}
	\left\|\frac{\mathrm{d}L^1_\sigma(p_c, x)}{\mathrm{dt}}\right\| \leq NM_1(MM_1 + 2K) \leq C_1^*, \nonumber
\end{equation}
for some global constant $C_1^* > 0$. Furthermore, since $\frac{\mathrm{d}}{\mathrm{dt}}\|L_\sigma^1\| = \frac{1}{2\|L_\sigma^1\|} \frac{\mathrm{d}}{\mathrm{dt}}\|L_\sigma^1\|^2 = \dot {L_\sigma^1}^\top \frac{L_\sigma^1 }{\|L_\sigma^1\|}$, we also have that
\begin{equation}
	\frac{\mathrm{d}}{\mathrm{dt}} \|L_\sigma^1\| \leq \left\| \frac{\mathrm{d}L^1_\sigma(p_c, x)}{\mathrm{dt}}\right\| \leq C_1^*. \nonumber
\end{equation}
Combining all the (conservative) calculated bounds, we can conclude that $	|w_d| \leq \left\| \frac{\mathrm{d}}{\mathrm{dt}} m_d \right\| \leq \frac{2C_1^* \|L_\sigma^1 \|}{ \|L_\sigma^1 \|^2} = \frac{2C_1^*}{\|L_\sigma^1 \|} \leq \frac{2C_1^*}{m\epsilon} = \Omega_d,$ where for the last inequality we have chosen $\kappa_\gamma \geq \kappa$ from Lemma \ref{lem: Lgrad} so that $x(t)$ is non-degenerated, and together with the assumption $\|\nabla\sigma(p_c(t))\| \geq \epsilon$ we can choose $m$ from Lemma \ref{lem: Lgrad} as well.
\end{proof}

\subsection{Proof of Proposition \ref{prop: gamma}} \label{proof: lem_gamma}

\label{proof: gamma}
\begin{proof}
Due to Lemma \ref{lem: Omegad} we have that if $\nabla\sigma(p_c(t))>\epsilon$, then $|\omega_d| \leq \Omega_d$ for all $t$, i.e., $|\omega_d| \leq \Omega_d, \forall t \in [0,T)$. Consequently, we consider $\kappa_\gamma \geq \kappa = \operatorname{max}\left\{\kappa^*, \frac{2\Omega_d}{\gamma},\frac{2}{T}\ln\left(\frac{\pi}{\gamma}\right)\right\}$, where $\kappa^*$ is chosen according to Lemma \ref{lem: x} so that $x(t)$ is non-degenerated. 

Given this choice of $\kappa_\gamma$, we firstly analyze the behavior of $\delta_i$. If $\delta_i \geq \gamma$ then $\dot\delta_i \leq -\kappa_\gamma\delta_i + \Omega_d \leq -\kappa_\gamma\delta_i + \kappa_\gamma \frac{\delta_i}{2}.$ Similarly, if $\delta_i \leq -\gamma$, we obtain $\dot\delta_i \geq -\kappa_\gamma\delta_i - \Omega_d \geq \kappa_\gamma \frac{\delta_i}{2}.$ Thus, the interval $I := (-\gamma, \gamma)$ is attractive. In particular, given the uniqueness of solutions, once $\delta_i(t)$ enters $I$, it remains as long as the conditions of Lemma \ref{lem: Omegad} are fulfilled; that is, until the first time $T^* \geq T$ such that $\|\nabla \sigma(p_c(T^*))\| \leq \epsilon$.
\end{proof}

\end{appendix}


\bibliographystyle{IEEEtran}
\bibliography{biblio}


\begin{IEEEbiography}[{\includegraphics[width=1in, height=1.25in, clip,keepaspectratio, trim={0cm 0cm 0cm 0cm}]{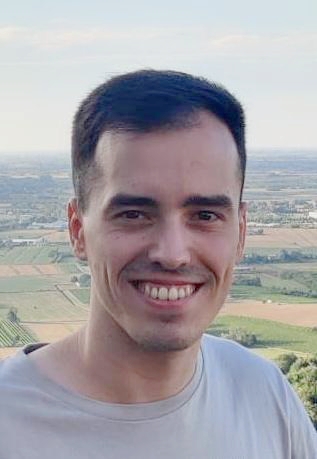}}]{Jesus Bautista Villar} (Student IEEE) received his B.S. degree in Physics from the Complutense University of Madrid, Spain, in 2022 and his M.S. degree in Data Science and Computer Engineering from the University of Granada, Spain, in 2023. He is currently pursuing a Ph.D. in Cognitive Systems and Robotics at the University of Granada, Spain, where his research focuses on multi-agent systems and the design of guidance navigation and control systems for autonomous aerial vehicles.
\end{IEEEbiography}

\begin{IEEEbiography}[{\includegraphics[width=1in, height=1.25in, clip,keepaspectratio, trim={2cm 0cm 2cm 0cm}]{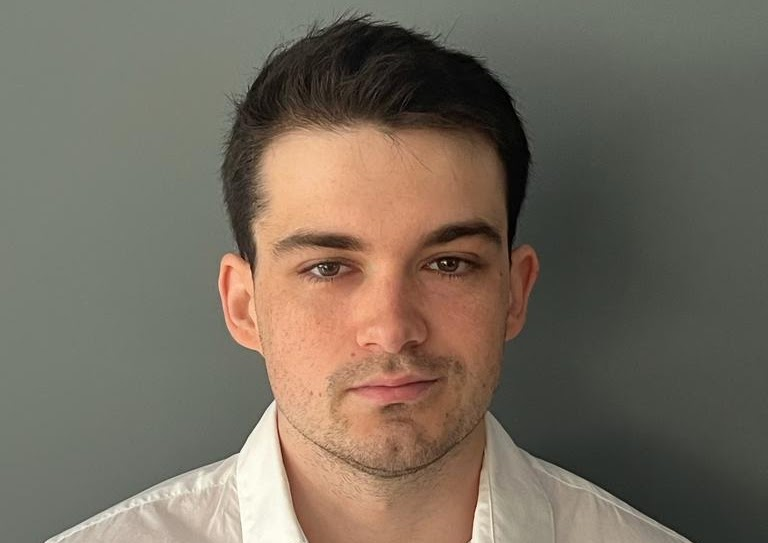}}]{Antonio Acuaviva} earned his B.S. degrees in Mathematics and Physics from Universidad Complutense de Madrid, Spain, in 2022. He then obtained his M.A.St. in Mathematics from the University of Cambridge, UK, in 2023, supported by a 'la Caixa' Fellowship. Currently, he is pursuing his Ph.D. in Mathematics at Lancaster University, UK, where his research focuses on functional analysis and Banach space theory.
\end{IEEEbiography}

\begin{IEEEbiography}[{\includegraphics[width=1in, height=1.25in, clip,keepaspectratio, trim={0cm 0cm 0cm 2cm}]{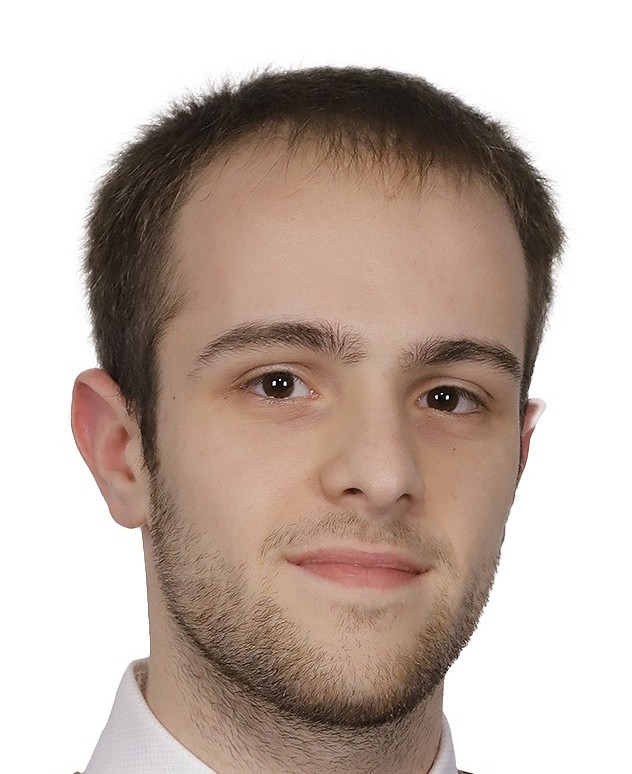}}]{Jose Hinojosa Hidalgo} completed his B.S. degree in Computer Engineering from the University of Granada, Spain, in 2024. He is currently pursuing a M.S. in Data Science and Computer Engineering from the University of Granada.
\end{IEEEbiography}

\begin{IEEEbiography}[{\includegraphics[width=1in,height=1.25in,clip,keepaspectratio]{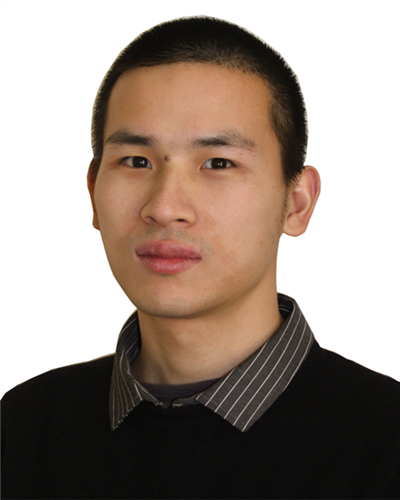}}]{Weijia Yao}  received the Ph.D. degree in systems and control theory from the University of Groningen, Groningen, The Netherlands. His research interests include nonlinear systems and control, robotics, and multiagent systems. He was a Finalist for the Best Conference Paper Award from ICRA in 2021.
\end{IEEEbiography}

\begin{IEEEbiography}[{\includegraphics[width=1in,height=1.25in,clip,keepaspectratio]{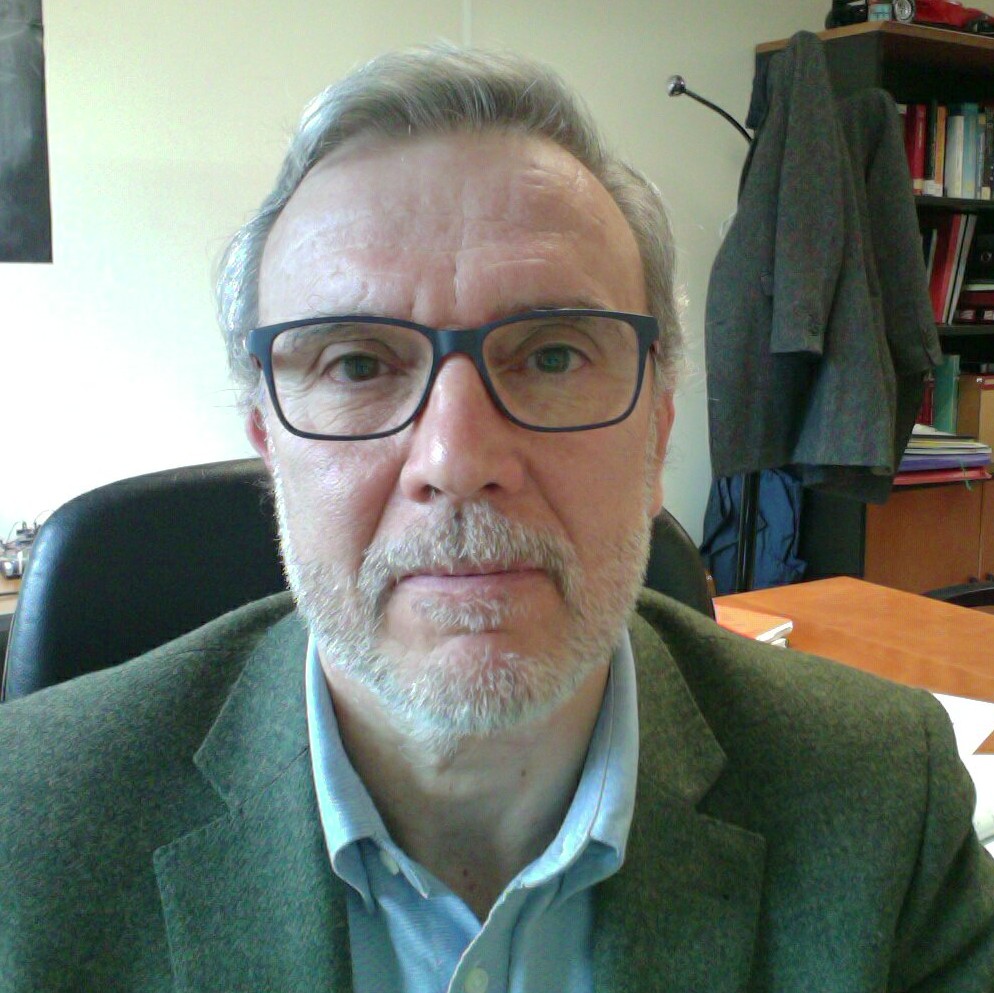}}]{Juan Jimenez} graduated in Physics from the Universidad Autónoma de Madrid (Spain) in 1986 and earned his Ph.D. in Systems Control in 1999 from the Universidad Nacional de Educación a Distancia (Spain). Since 2015, he has been an Associate Professor in the Department of Computer Architecture, Systems Engineering, and Automation at the Universidad Complutense de Madrid. His research interests include distributed control and cooperative control in multi-agent systems, with a particular focus on applications to autonomous vehicles.
\end{IEEEbiography}

\begin{IEEEbiography}[{\includegraphics[width=1in,height=1.25in,clip,keepaspectratio]{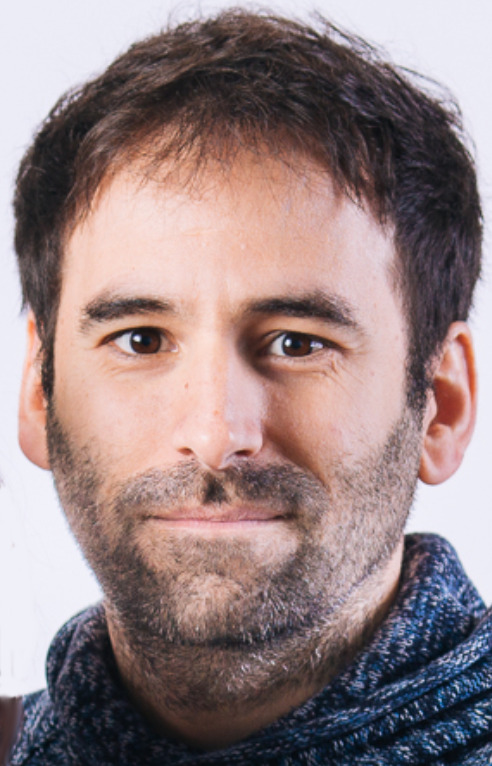}}]{Hector Garcia de Marina} (Member IEEE) received the Ph.D. degree in systems and control from the University of Groningen, The Netherlands, in 2016. He was a Postdoctoral Research Associate with the Ecole Nationale de l’Aviation Civile, Toulouse, France, and an Assistant Professor with the Unmanned Aerial Systems Center, University of Southern Denmark, Odense, Denmark. Since 2022, he has been a Ramón y Cajal Researcher with the Department of Computer Engineering, Automation and Robotics, and with CITIC, Universidad de Granada, Spain. He is the recipient of an ERC Starting Grant and an Associate Editor for IEEE Transactions on Robotics. His current research interests include multiagent systems and the design of guidance navigation and control systems for autonomous vehicles.
\end{IEEEbiography}

\end{document}